\documentclass[10pt,twocolumn,letterpaper]{article}
\usepackage{cvpr}

\cvprfinalcopy %

\def\cvprPaperID{8008} %

\ifcvprfinal\fi

\makeatletter

\def\eg{\emph{e.g}\onedot} 
\def\ie{\emph{i.e}\onedot} 
 
\def\etc{\emph{etc}\onedot} 
\def\wrt{w.r.t\onedot}

\makeatother

\usepackage{bm}
\usepackage[utf8]{inputenc}
\usepackage{times}
\usepackage{epsfig}
\usepackage{graphicx}
\usepackage{amsmath,mathtools}
\usepackage{amssymb}
\usepackage{amsthm}
\usepackage{bm}
\usepackage{enumitem}
\usepackage{mathtools}
\usepackage{multirow}
\usepackage[htt]{hyphenat}
\usepackage{fontawesome}
\usepackage{pifont}
\usepackage{color}
\usepackage{float}
\usepackage{xcolor}
\usepackage{subfig}
\usepackage[algo2e,inoutnumbered,linesnumbered,algoruled,vlined]{algorithm2e}
\usepackage{algpseudocode}
\usepackage{booktabs}
\usepackage{tabularx}
\usepackage{wrapfig}
\usepackage{appendix}
\usepackage{url}
\usepackage{bbm}
\usepackage{leftidx}
\usepackage{dcolumn}
\usepackage{gensymb}
\usepackage{tikz}
\usepackage{overpic}
\usepackage[pagebackref=true,breaklinks=true,colorlinks,bookmarks=false]{hyperref}
\newcommand{\name}{CIRCLE}
\newcommand{\netname}{CircNet}

\renewcommand{\P}{\mathcal{P}}  % Point cloud set
\newcommand{\p}{\bm{p}}       % a point
\newcommand{\normal}{\bm{n}}  % normal on point
\newcommand{\lvec}{\bm{l}}
\renewcommand{\c}{\bm{c}}    % center of a volume
\newcommand{\V}{\mathcal{V}}

\newcommand{\depth}{\mathcal{D}}
\newcommand{\pose}{\mathbf{T}}

\newcommand{\rayc}{\bm{o}}
\newcommand{\rayd}{\bm{d}}

\newcommand{\encnet}{\phi_\mathrm{E}}
\newcommand{\unet}{\phi_\mathrm{U}}
\newcommand{\decnet}{\phi_\mathrm{D}}

\newcommand{\gt}{\mathrm{gt}}
\newcommand{\R}{\mathbb{R}}
\newcommand{\SE}{\mathbb{SE}}

\newcommand{\eqgap}{\hspace{0.2em}}       % a point
\usepackage{cleveref}

\crefname{section}{Sec.}{Sec.}
\crefname{subsection}{Sec.}{Sec.}
\crefname{conj}{Conj.}{Conj.}

\Crefname{assumption}{\textbf{H}\hspace{-3pt}}{\textbf{H}\hspace{-3pt}}
\crefname{assumption}{\textbf{H}}{\textbf{H}}

\crefname{algorithm}{\text{Alg.}}{\text{Alg.}}
\crefname{assumption}{\textbf{H}}{\textbf{H}}
\crefname{equation}{\text{Eq}}{\text{Eq}}
\crefname{definition}{\text{Dfn.}}{\text{Dfn.}}
\crefname{lemma}{\text{Lemma}}{\text{Lemma}}
\crefname{dfn}{\text{Dfn.}}{\text{Dfn.}}
\crefname{thm}{\text{Thm.}}{\text{Thm.}}
\crefname{tab}{\text{Tab.}}{\text{Tab.}}
\crefname{fig}{\text{Fig.}}{\text{Fig.}}
\crefname{table}{\text{Tab.}}{\text{Tab.}}
\crefname{figure}{\text{Fig.}}{\text{Fig.}}
\crefname{section}{\text{Sec.}}{\text{Sec.}}

\newcommand{\parahead}[1]{\vspace{1.5mm}\noindent\textbf{#1.}\ }

\newcommand{\chx}[1]{{#1}}

\title{\name{}: Convolutional Implicit Reconstruction and Completion for \\Large-scale Indoor Scene}
\author{Haoxiang Chen \quad Jiahui Huang \quad Tai-Jiang Mu \quad Shi-Min Hu\thanks{corresponding author.}\\
BNRist, Department of Computer Science and Technology, Tsinghua University,
Beijing
}

\begin{document}

\twocolumn[{%
\renewcommand\twocolumn[1][]{#1}%
\maketitle
\begin{center}
    \vspace{-1em}
    \centering
    \captionsetup{type=figure}
    \includegraphics[width=\linewidth]{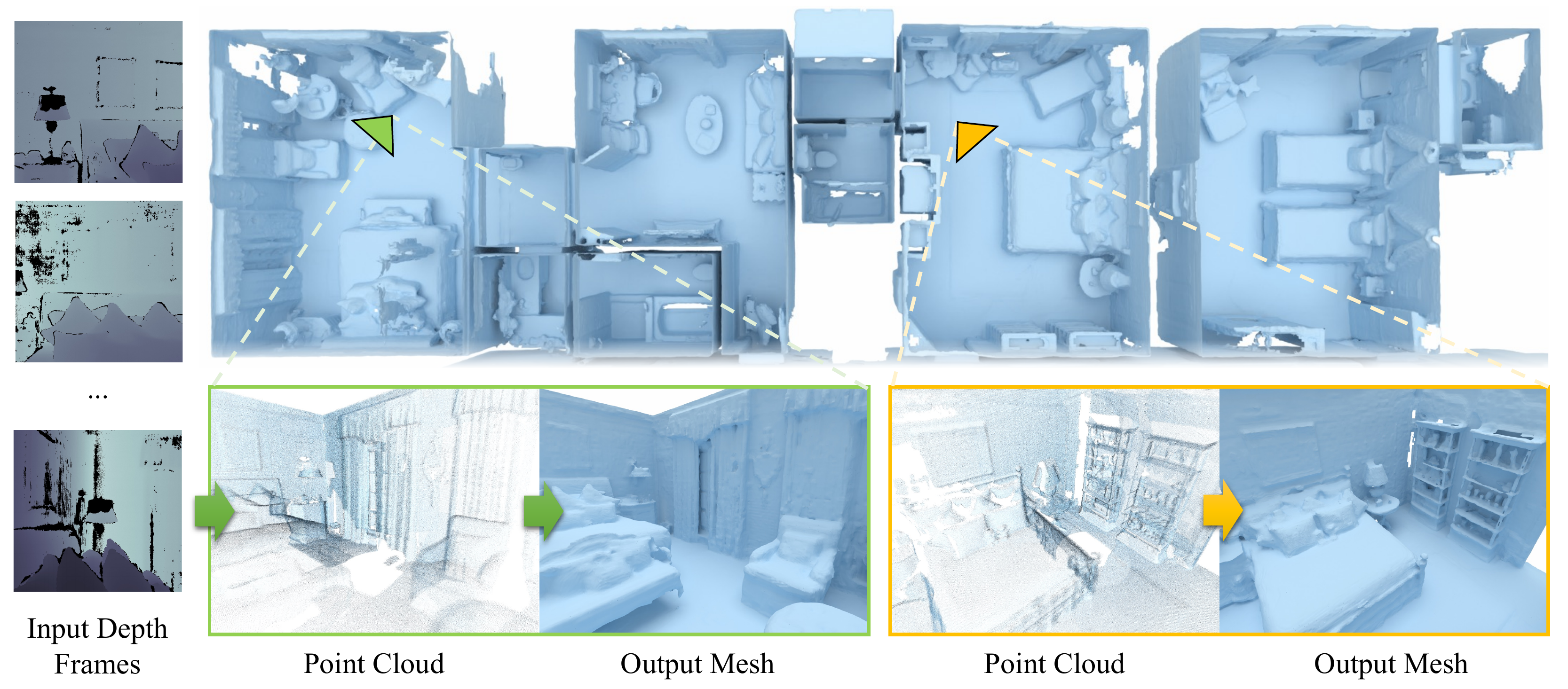}
    \vspace{-1.5em}   
    \captionof{figure}{\textbf{\name.} Given a sequence of depth images, with pose, corrupted by noise and missing data,
      our framework models the scene geometry and  contextual information with
    a fully-convolutional neural network, generating a high-quality, complete mesh for the underlying scene, which is represented using
    local implicit grid. Benefiting from sparsity, our method is fast and accurate: the inference time for this scene is only 17s, 10$\times$ faster than the method in~\cite{peng2020convolutional}.}
\label{fig:teaser}
\vspace{1em}
\end{center}%
}]

\maketitle

\begin{abstract}
We present \name{}, a framework for large-scale scene completion and geometric refinement based on local implicit signed distance functions.
It is based on an end-to-end sparse convolutional network, \netname{}, that jointly models local geometric details and global scene structural contexts, allowing it to preserve fine-grained object detail while recovering missing regions commonly arising in traditional 3D scene data.
A novel differentiable rendering module enables test-time refinement for better reconstruction quality.
Extensive experiments on both real-world and synthetic %various 
datasets show that our concise framework is  efficient and effective, achieving  better reconstruction quality than the closest competitor while being 10--50 $\times$ faster.
\end{abstract}

\section{Introduction}
\vspace{-0.3em}
In recent years, 3D reconstruction from RGB-D camera data has been widely explored thanks to its ease of acquisition with many applications in robotic perception, virtual reality,  games, \etc.
It is well-accepted that an ideal reconstruction algorithm should be capable of simultaneously (i) restoring fine-grained geometric detail in the target scene, (ii) handling
 large scenes efficiently, and (iii) completing missing regions of the scene.
Additionally, the underlying 3D representation should be flexible enough to allow further optimization of geometric quality.

However, traditional algorithms along with their accompanying representations fail to effectively fulfil the above requirements.
For instance, methods using the
\emph{truncated signed distance function} (TSDF)~\cite{curless1996volumetric,newcombe2011kinectfusion} are hampered by limited voxel resolution and lack robustness to noisy data.
Surfels~\cite{whelan2015elasticfusion} offer more flexibility by treating the 3D scene %reconstruction
as unstructured points, but maintaining  correct topology is challenging.
Furthermore, such methods as these cannot fill in missing geometry in the scene, which is common in practice due to sensor limitations, incomplete coverage of the scanning trajectory, or unreachable areas.

The recent introduction of deep implicit representations~\cite{park2019deepsdf,mescheder2019occupancy,chen2021learning} has enabled a plethora of research directions for 2D and 3D data processing.
Parameterized by a neural network, implicit functions are inherently continuous and differentiable. % due to the nature of the neural networks. 
Notably, in the field of 3D reconstruction, various works~\cite{chen2020bsp,chibane2020implicit,Genova_2020_CVPR,Genova_2019_ICCV,oechsle2019texture} have already demonstrate their ability to learn \emph{object-level} geometric priors from shape repositories.
However, when applied to large-scale scenes, the above methods are typically impractical, fro three reasons. Firstly,
scene structures are substantially more complicated than a single object. A typical end-to-end, optimization-free, framework is weak at capturing the entangled geometric priors of cluttered regions.
Secondly, while other work exists~\cite{Jiang_2020_CVPR,sitzmann2019siren,takikawa2021neural,azinovic2021neural} that overfits the scene geometry, to avoid the necessity of prior learning,
it usually involves costly optimization procedures.
Thirdly, some efforts~\cite{peng2020convolutional,huang2021di,Sucar_2021_ICCV} have been made to reconstruct  scenes in real-time with deep implicit functions, but they do so at the cost of low reconstruction quality.

To tackle these issues, we introduce the \name{} framework, as shown in \cref{fig:teaser}. It employs a novel \netname{}, short for fully-convolutional implicit network for reconstruction and completion of large-scale indoor 3D scenes from partial point clouds. It is capable of both preserving scene geometric details and completing missing regions of the scene in a semantically-meaningful way.
Specifically, we adopt local implicit grid to represent  local details of the overall scene, and learn global contextual information for scene completion via a sparse U-Net.
%\tjc{need some more sentences to explain how your method achieving the aforementioned capability}
Our network is also efficient, in that it encodes and decodes the sparsity pattern of the scene geometry by learning, and only non-empty portions need to be evaluated.
Furthermore, we provide a fast and novel differentiable rendering approach tailored for refining our output representation, which can greatly improve the geometric quality during inferencing to provide resilience in the face of  errors in the raw input. %\tjc{what motivates such a DR branch}% refinement.
Extensive experiments using various datasets demonstrate the effectiveness of our framework, which sets a new state-of-the-art for scene reconstruction and completion. In benchmarks it is 10--50 $\times$ faster than previous methods.

\section{Related Work}

\parahead{Scene Reconstruction}%
Building a high-quality and coherent scene-level reconstruction is challenging due to  noise, occlusion and missing data inherent in 3D data acquisition sensors.
While traditional methods~\cite{newcombe2011kinectfusion,whelan2015elasticfusion,dai2017bundlefusion,oleynikova2017voxblox} incrementally fuse input depth observations using a moving average~\cite{curless1996volumetric}, learning methods~\cite{weder2020routedfusion,weder2021neuralfusion} can further reduce  noise using data-driven geometric biases.
The recent trend of using implicit neural representations, such as DI-Fusion~\cite{huang2021di} and its successors~\cite{sucar2021imap,bovzivc2021transformerfusion}, either uses localized priors or the continuous nature of a globally-supported network function.
In comparison, our method can not only accurately recover detailed scene geometry, but also rebuild missing parts via global structural reasoning based on learning.

\parahead{Scene Completion}%
The main challenge in scene completion is to fill missing regions with data that are semantically coherent with the existing content.
\cite{song2018im2pano3d} casts the problem in terms of panoramic image completion but important geometric details are significantly missing.
\cite{dai2018scancomplete} first brings the aid of semantic segmentation to the completion problem in the 3D domain.
%%%RRM There is a lot of earlier work on 3D scene completion
%%%RRM e.g. see this paper from 2205
%%%RRM https://breckon.org/toby/publications/papers/breckon05amodal.pdf
Subsequent lines of work~\cite{dai2020sg,dai2021spsg} tackle the problems of geometric sparsity and color generation.
We note that many end-to-end frameworks~\cite{peng2020convolutional,azinovic2021neural} using implicit representations also provide decent scene extrapolation due to the continuous nature of networks, even though they are not specifically designed for this task.

\parahead{Differentiable Rendering}%
The technique of differentiating the rendering process bridges the gap between 3D geometry and 2D observations of it by allowing for end-to-end optimization directly from captured raw sensor data, which was first applied to triangular meshes~\cite{liu2019soft,kato2018neural} and later to implicit fields~\cite{liu2020dist,niemeyer2020differentiable}.
The prevalence of NeRF~\cite{mildenhall2020nerf} motivates many studies to improve rendering efficiency and fitting speed, either through localized structures~\cite{liu2020neural}, level-of-detail rendering~\cite{takikawa2021neural}, caching~\cite{yu2021plenoctrees}, or multi-view stereo~\cite{yariv2020multiview,rosu2021neuralmvs}.
In conjunction with our novel local implicit representation, we devise a new differentiable rendering approach can rapidly and effectively refine detail geometries of the reconstructed scene during inferencing.

\begin{figure*}
    \centering
    \includegraphics[width=\linewidth]{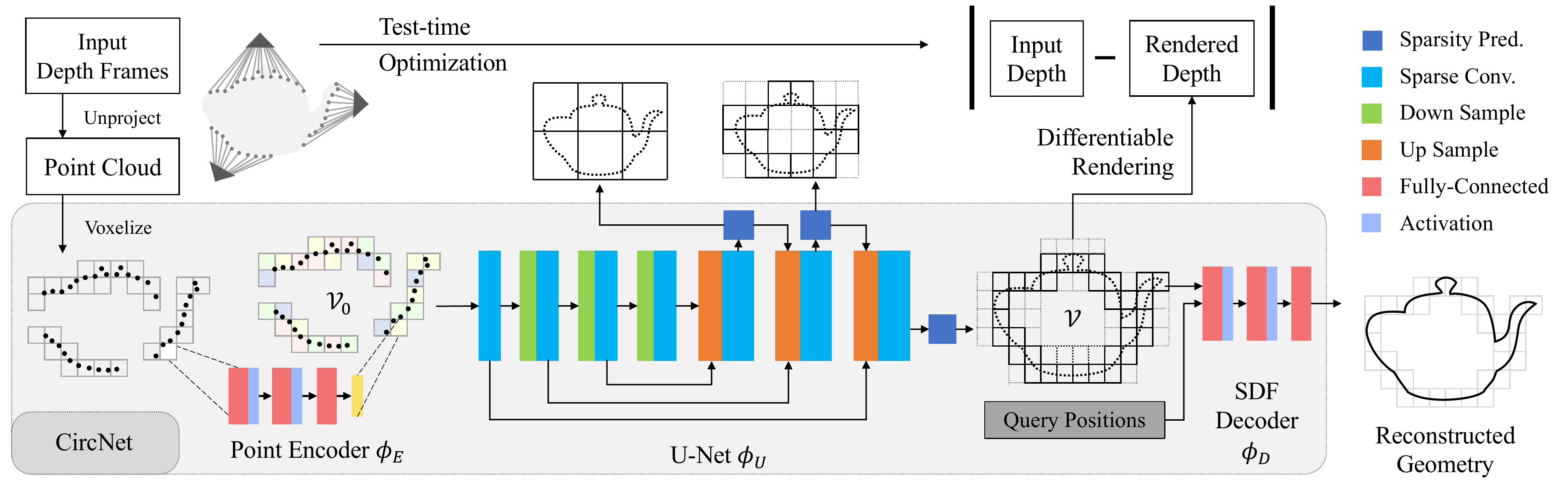}
    \caption{\textbf{Pipeline.} We first voxelize the accumulated unprojected points from the \chx{input posed depth frames} into sparse grid. The feature volume is then passed through \netname{}, which comprises 3 neural networks: $\encnet$, $\unet$, and $\decnet$. The output reconstructed geometry is an implicit completed surface. Inference-time refinement is enabled by our differentiable rendering algorithm; it optimizes both  scene geometry and  camera pose, leading to better and more complete reconstruction.}
    \label{fig:overview}
    \vspace{-1em}
\end{figure*}

\section{\name{}: {Convolutional Implicit Scene Reconstruction and Completion}}
% method overview

\parahead{Problem Formulation}%
The input to our method is a sequence of depth frames with pose $\{ \depth_t, \pose_t \}_{t=1}^T$, with $\depth_t \in \R^{W \times H}$ and $\pose_t \in \SE(3)$ being the depth image and the %given
6-DoF camera pose, respectively.
Our goal is to build a high-quality and complete 3D reconstruction of the scene, represented using $M$ %in a 
local sparse implicit voxel grid $\V = \{(\c_m,\lvec_m)\}_{m=1}^M$ %containing only the $M$ voxels 
that contain
the surface of the scene geometry. Here, $\c_m \in \R^3$ is %the position of 
the voxel coordinate and $\lvec_m \in \R^L$ is the latent vector describing the local voxel grid's geometry, from which we can decode the signed distance values of the full scene and finally extract the mesh. The size %side length 
of each voxel is $b\times b\times b$.

% Due to the nature of depth scans, $\P$ from real-world data can suffer from substantial noise and partiality.

\parahead{Overview}%
As \cref{fig:overview} shows, we first unproject all the depths $\depth_t$ under the given poses $\pose_t$ to obtain an accumulated point cloud $\P = \{(\p_i,\normal_i)\}_{i=1}^N$ where $\p_i \in \R^3$ and $\normal_i \in \R^3$ are  point positions and their estimated normals, using~\cite{newcombe2011kinectfusion}.
$\P$ is then voxelized into initial sparse 3D grid and processed by  \netname{} (see \cref{med:net}).
Being aware of both global scene structure and local geometric details, \netname{} simultaneously refines the voxelized points and adds additional points with a point encoder $\encnet$ and U-Net $\unet$, and produces $\V$ defining the latent vector of local implicit geometry, which is then decoded to TSDF values %parametrized 
by a multi-layer perceptron (MLP) $\decnet$.
One can later extract the mesh using marching cubes~\cite{lorensen1987marching} from these TSDF values.
Moreover, the reconstructed geometry can be further optimized during inferencing time via a novel differentiable rendering scheme described in \cref{med:dr}, to refine both the scene geometry and the camera pose. 
Detailed loss functions for the training procedure and inference-time refinement are discussed in \cref{med:train}.

\subsection{\netname{} Architecture}
\label{med:net}
Given the unprojected point cloud $\P$ from the input views,  \netname{} sequentially applies three trainable components: a point encoder network $\encnet$, a U-Net $\unet$, and an SDF decoder $\decnet$ to produce an implicit representation of the underlying scene. We now describe these in turn.

\parahead{Point Encoder}%
We first split the input point cloud into multiple voxel grid.
For point $\p_i$, the index of its corresponding voxel $m_i$ is determined by $m$ satisfying $\p_i \in [\c_m , \c_m + b)$.
We define the local coordinates of $\p_i$ within its voxel as $\p_i^l = (\p_i - \c_m) / b \in [0,1]^3$.
Next, for each voxel $m$, we feed all  local coordinates of  points within the voxel, along with their normals: $\{ (\p_i^l, \normal_i) \in \R^6 \eqgap | \eqgap m_i = m \}$ into a point encoder $\encnet$.
This uses a basic PointNet~\cite{qi2017pointnet} structure by first  mapping all the input features into $L$-dimensions with a shared MLP and then aggregating the features via mean pooling. 
The resulting sparse feature voxel grid is denoted $\V_0$.

% Additionally, toward adapting scan data such as RGB-D scans, the feature locates on the center of voxels, in stead of locating on the corner of voxels. This allows an incremental point encoding as follows. We store all voxels with its latent vectors $\bm{l}_m$ and $\bm{w}_m$ the number of points encoded in. When new observation coming, the observation points can be encode to $\bm{l}_m^t$ and its number of points $\bm{w}_m^t$. the latent vectors can be updated as:
% \begin{equation}
% \label{latent_update}
% \bm{l}_m \leftarrow \frac{\bm{l}_m\bm{w}_m + \bm{l}_m^t\bm{w}_m^t}{\bm{w}_m + \bm{w}_m^t},
% \bm{w}_m \leftarrow \bm{w}_m + \bm{w}_m^t
% \end{equation}
% It's clearly that this incremental update equation equivalent to encode all point at the same time.

\parahead{U-Net}%
The goal of the U-Net $\unet$ in this step is to complete and refine the reconstruction from $\V_0$ into $\V$. % through simultaneous densifying the geometry %modifying the sparsity 
% and refining the features in the voxel grid.
This is done by propagating contextual features in the hierarchical U-Net structure with a large receptive field.
A trivial implementation falls back to a dense convolution that generates a dense feature grid even if many voxels are actually empty.
Due to the sparse nature of the geometry, we instead use submanifold sparse convolution~\cite{SubmanifoldSparseConvNet} for our convolution layer.
% For the decoder branch, 
For the decoder branch, inspired by~\cite{wang2020deep}, we append a \emph{sparsity prediction} module to each layer of the decoder. 
This module is instantiated with a shared MLP applied to each voxel and predicts the confidence of the current voxel containing true surfaces; voxels with scores lower than 0.5 are pruned. 
Accordingly, usual skip connections are replaced by sparsity-guided skip connections: connections are only added for  voxels predicted to be non-empty.
Apart from the efficiency gain, this design also eases network training by obviating the need to model the full geometry of empty regions.

% By using such operations, the u-net together with sdf-decoder don't need to learn representation of a grid without surfaces, which makes network easier for training. 
%In addition, for smoothing surface across feature grid boundaries, the output feature locate on the corner of grids, enabling trilinear interpolation for continue sdf prediction step.

\parahead{SDF Decoder}%
To recover the final scene geometry, we traverse all  points $\p$ in the non-empty regions of $\V$ and learn signed distance values  using an implicit decoder instantiated with an MLP $\decnet: (\p^l, \hat{\lvec}) \in \R^{3+L} \mapsto [-1,1]$, where $\p^l$ is the local coordinate of $\p$  and $\hat{\lvec}$ is the interpolated feature taken from $\V$.
To achieve smooth geometric interpolation across  voxel boundaries, we apply an additional $2\times 2 \times 2$ convolution over $\V$ to propagate the features stored at voxel \emph{centers} to voxel \emph{corners}, obtaining $\{\lvec'_m \}$.
The input feature $\hat{\lvec}$ can  then be trilinearly interpolated $\psi(\cdot)$ from the features stored at its 8 nearest voxel corners: $\hat{\lvec} = \psi(\p^l, \{ \lvec_{(1)}', \dots, \lvec_{(8)}'\} )$.
% Denote the feature vector of point $\p$ and feature grids $\V$ as $\psi(p,\mathcal{V})$, we predict sdf value through a small shared-MLP.

% \begin{equation}
% \label{sdf_predict}
% \decnet(\p, \psi(\p,\mathcal{V})) \rightarrow 
% \end{equation}

\subsection{Differentiable Local Implicit Rendering}
\label{med:dr}
\begin{figure}
\includegraphics[width=\linewidth]{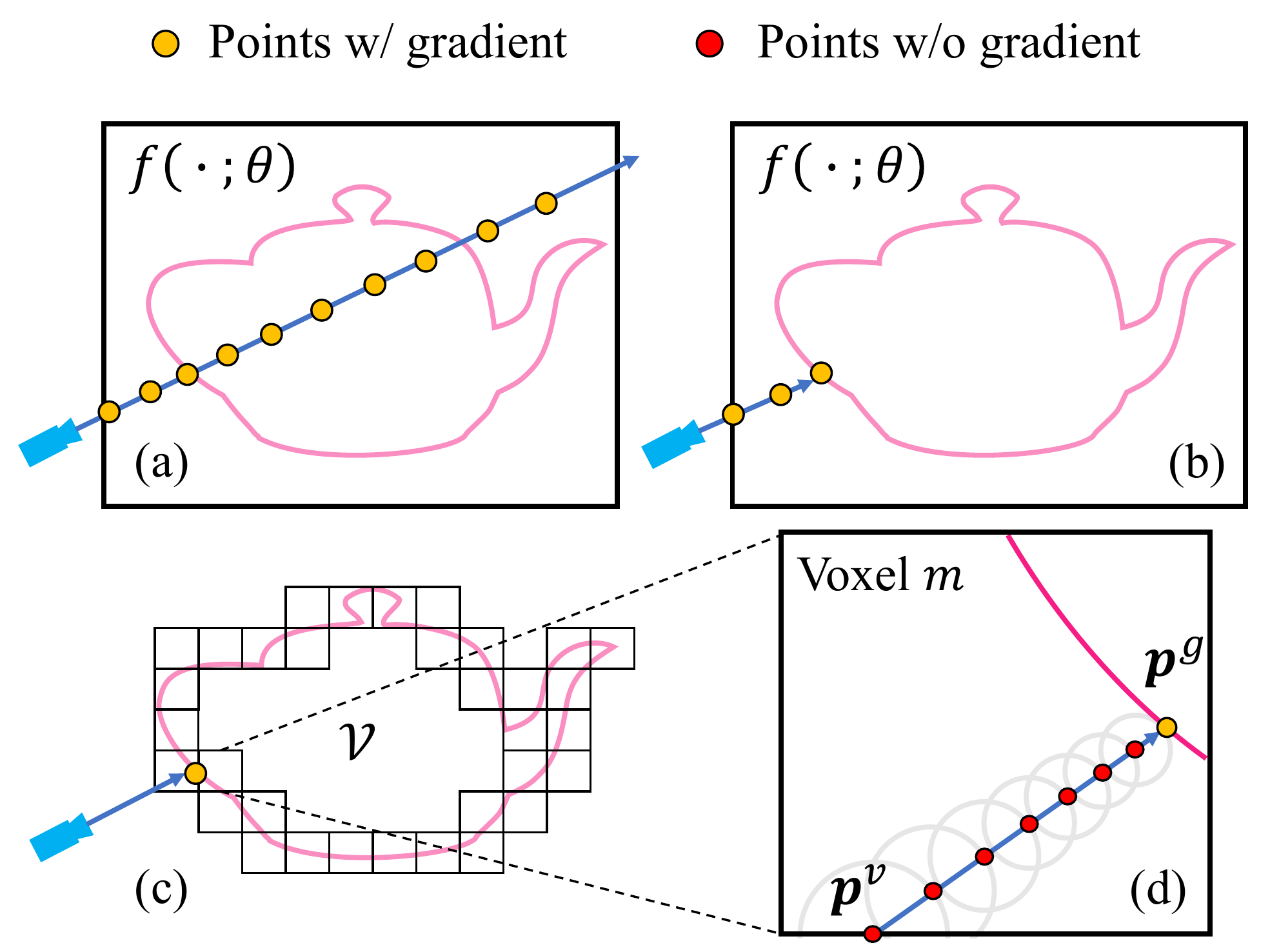}
\caption{\textbf{Rendering strategies.} Rendering a globally-supported implicit representation with (a) uniform query points~\cite{mildenhall2020nerf} and (b) differentiable sphere tracing~\cite{liu2020dist}. (c,d) Our approach with both sphere tracing and implicit differentiation, where $\p^i=\rayc + t^i \rayd, i \in \{v,g\}$.}
\label{fig_dr}
\end{figure}

Despite the good-quality, end-to-end reconstruction provided by \netname{}, some desired geometric details can be lost.
The reasons are two-fold.
Firstly, real-world depth captures usually suffer from noisy pose and sensor limitations, resulting in erroneous reconstruction and severe missing regions. 
Secondly, a simple feed-forward network trained on large-scale datasets can underfit geometric features or generate excessive contents~\cite{park2019deepsdf,mescheder2019occupancy}. 
Noting these issues, we propose a novel differentiable renderer for our implicit representation, allowing for effective differentiation through both geometry and  camera pose.
Specifically, for each pixel to be rendered, we emit a ray with an origin $\rayc$ and a unit direction $\rayd$, and compute the depth of the intersection $t$ so that the intersection point is $\p = \rayc + t \rayd$, and forward and backward passes are defined as follows:

\parahead{Forward Pass}
The forward pass is composed of two steps as shown in \cref{fig_dr} (c--d):
\begin{enumerate}[topsep=0pt,leftmargin=*]
\setlength{\itemsep}{0pt}
\setlength{\parskip}{0pt}
    \item \textbf{Voxel-level Intersection.} As the sparsity prediction modules from the different layers of our U-Net decoder naturally form an \emph{octree} structure thanks to the upsampling operator, we can use any existing ray-octree intersection algorithm for this step. In our implementation, we choose the fast algorithm in~\cite{takikawa2021neural} that generates a list of intersection pairs $\{(t^v,m^v)\}$, where $t^v$ is the depth and $m^v$ is the voxel index of the intersection.
    \item \textbf{Geometry-level Intersection.} The sphere tracing algorithm~\cite{hart1996sphere} is applied for each intersecting voxel $m^v$, starting from $\rayc + t^v \rayd$ and ending at $\p^g = \rayc + t^g \rayd$ that hits the surface. Note that only the smallest $t^g$ among all the voxels is returned as the final depth $t$ due to occlusion.
\end{enumerate}

\parahead{Backward Pass}
For clarity, %To facilitate discussion, 
we abstract our full \netname{} as an implicit network $f(\p ; \theta)$ whose inputs are the position $\p$ and the intermediate features or network parameters $\theta$, 
and the output is the signed distance value. % at our disposal.
We wish to compute the first-order derivative of the depth $t$ \wrt $\theta$ as well as the camera ray $\rayc$ and $\rayd$ for optimization.
Inspired by~\cite{yariv2020multiview}, we employ the fact that $f(\rayc + t\rayd; \theta) \equiv 0$ and use  implicit differentiation to obtain:
\begin{align}
\label{ep:dr}
    \frac{\partial t}{\partial \theta} = -\gamma \frac{\partial{f}}{\partial \theta}, \quad
    \frac{\partial t}{\partial \rayc} = -\gamma \frac{\partial f}{\partial \p}, \quad
    \frac{\partial t}{\partial \rayd} = -\gamma t  \frac{\partial f}{\partial \p},
\end{align}
where $\gamma = \langle \rayd, {\partial f}/{\partial \p} \rangle ^{-1}$ is a scalar, $\langle\cdot,\cdot\rangle$ denotes vector inner product, and other derivatives related to $f$ can be efficiently evaluated using reverse-mode back-propagation.
Empirically, we observe that full gradient-based optimization %computation 
over all network parameters fails to converge. %lead to diverged optimization result.
Hence we choose to  only optimize the latent vectors in $\V$: $\theta = \{ \lvec_m \}$, and fix all other parts of the networks.

% The ray marching step compute intersect point $p = \mathbf{c} + t\mathbf{d}$, we should compute it in a differentiable way.
% We construct first order approximation of feature volumn $\mathcal{V}$, ray direction $\mathbf{d}$ and ray origin $\mathbf{c}$. 
% Denote $p_0=\mathbf{c}_0 + t_0\mathbf{d}_0, \mathbf{x}_0 = \psi(p_0,\mathcal{V}_0)$ and $\mathbf{x} = \psi(p,\mathcal{V})$. 
% The surface intersection can be expressed as:
% \begin{equation}
% \label{diff_intersect}
% t(\mathbf{c},\mathbf{d},\mathcal{V}) = t_0 - \frac{\phi_D(p,\mathbf{x}) - \phi_D(p_0,\mathbf{x}_0)}{\nabla_p(p_0,\mathbf{x}_0)\cdot\mathbf{d}_0}
% \end{equation}

\parahead{Discussion}
A comparison between our method and previous approaches is shown in \cref{fig_dr}.
Methods similar to, \eg, NeRF~\cite{mildenhall2020nerf} exhaustively query all  points along the ray; most of the unnecessary computations far away from the surface can be saved with sphere tracing~\cite{hart1996sphere,liu2020dist}. 
Our use of \emph{localized} grid further speed up the process thanks to the explicit ray-voxel intersection step that greatly reduces the number of steps in tracing.
Nevertheless, a naive implementation of the backward pass requires unrolling the tracing steps, leading to inaccurate gradients.
We for the first time marry the merits of implicit differentiation, originally designed for global representations~\cite{yariv2020multiview}, with our local feature grid, so that only the intersection points need to be stored in the computation graph, leading to a fast, stable, accurate and memory-efficient method for both  forward and backward passes.
Experiments verifying our design choices are shown in \cref{ablation}.

% The differentiable renderers applied by global implicit representations like DIST~\cite{liu2020dist},NeRF\cite{mildenhall2020nerf} are usually suffering from the speed, volume rendering needs query lots of  points that exactly far away from the surface. 
% SDF-based differentiable rendering avoids query points behind the hit surface, however they must query points step by step and spend long time if there are many steps. 
% Based on local implicit representations, SDF-based needs much smaller number of steps and there are two alternative method: 
% 1) differentiable ray marching, similar to DIST, compute gradient of all queried points and simply sum up them, called Local-DIST in following part. 2) differentiable ray intersection, inspired by IDR\cite{yariv2020multiview}, implicit differentiate the intersection point and only compute the gradient of the hit point. Local-IDR is faster and more memory efficient than Local-DIST because it only the intersection points are in the compute graph, and we will demonstrate that Local-IDR is more accurate on the derivative of the poses in .

\subsection{Loss Functions}
\label{med:train}

\parahead{\netname{} Loss Function}%
The three networks $\encnet$, $\unet$ and $\decnet$ are jointly trained in an end-to-end manner, using the following loss function:
\begin{equation}
\label{total_loss}
\mathcal{L} = \mathcal{L}_\mathrm{sdf} + \alpha \mathcal{L}_\mathrm{norm} + \beta \mathcal{L}_\mathrm{struct} + \delta \sum_{m=1}^M \lVert \lvec_m \rVert,
\end{equation}
where $\lVert \cdot \rVert$ is the vector norm. $\mathcal{L}_\mathrm{sdf}$ is the data term defined as the L1 distance between the predicted signed distance from the decoder $\decnet(\p^l, \hat{\lvec})$ and the ground-truth values $s^{\gt}(\p)$:
\begin{equation}
\label{sdf_loss}
\mathcal{L}_\mathrm{sdf} = \int_{\Omega_u\cup\Omega_n} {|\decnet (\p^l,\hat{\lvec}) - s^\gt (\p)|} \eqgap \mathrm{d} \p.
\end{equation}

Here $\Omega_u$ denotes the occupied region of the voxels $\V$ while $\Omega_n$ is a narrow band region near the surface. 
%%%RRM Surely this should depend on the scale of the input data.
%%%RRM What if my input points have scales of metres, e.g. from an aerial scan?
% Notably, we refrain from applying a truncation to the signed distance, mainly due to the homogeneity exposed in the training samples originating from the small voxel size we use, which is detrimental for training convergence and generalizability of the learned implict field.
The normal of the predicted geometry, computed as $\nabla_{\p}\decnet$, is constrained by the normal loss:
\begin{align}
\label{eq:normal_loss}
\mathcal{L}_\mathrm{norm} = & \int_{\Omega_u\cup\Omega_n} \big| \| \nabla_{\p}\decnet\| - 1 \big| \eqgap \mathrm{d} \p \eqgap + \nonumber\\
& \int_{\Omega_n} \big( 1 - \langle \nabla_{\p} \decnet, \normal^{\gt} (\p) \rangle \big) \eqgap \mathrm{d} \p,
\end{align}
where the first term enforces the eikonal equation of the signed distance field while the second term minimizes the angle between predicted normal and  ground-truth normal $\normal^\gt$.

$\mathcal{L}_\mathrm{struct}$ uses cross-entropy loss to supervise the sparsity prediction module for each layer in the decoder branch of $\unet$.
Specifically, we obtain the ground-truth sparsity pattern of the target geometry at multiple resolutions in accordance with the output sparsity map from the U-Net, and directly supervise the predicted confidence score.
% Denote the layer number of $\phi_U$ as $L$, then we construct a $L$-half octree of the ground-truth points. 
% The node status of octree can directly supervise the predicted result. 
During training, we use the ground-truth sparsity map instead of the predicted one for the skip-connections and pruning of the next layer. 
% The last term in \cref{total_loss} regularizes the norm of the feature vector in output feature volumn $\mathcal{V}$ with an L2 loss.

\parahead{Inference-time Refinement}%
During inferencing, our differentiable rendering module is applied to refine the predicted geometry and the camera poses. For each depth image $\depth_t$ and its pose $\pose_t$, we can render a depth image as $\depth'_t(\pose_t,\theta) \in \R^{W\times H}$, whose pixels are the depths $\{t\}$ from \cref{med:dr}. 
By minimizing the error between the rendered depth and the observed depth, we can jointly optimize the quality of geometry and input poses:  
%Here we do not direct optimize latent vector generated by u-net, because through a u-net gradient can be propagate to all the grid in the receptive field,  
\begin{equation}
\label{render_optimize}
\min_{ \theta, \{\delta\pose_t \}} \eqgap \eqgap \sum_{t=1}^T \big| \depth_t - \depth'_t(\delta\pose_t\pose_t,\theta)) \big|,
\end{equation}
where %where for poses
we optimize an increment to pose $\delta\pose_t$ instead of $\pose_t$ itself, for better convergence.

\section{Experiments}

\subsection{Dataset and Settings}

\parahead{Datasets}%
The main dataset used to evaluate our framework is {N-Matterport3D}.
Adapted from~\cite{Matterport3D}, this dataset contains 1788 + 394 (for training / validation and testing respectively) scans of rooms from 90 buildings captured by a Matterport Pro Camera.
We follow the self-supervised setting from~\cite{dai2020sg} by randomly sampling 50\% of the frames to generate an incomplete version of each room and supervise our method with a complete version reconstructed from all frames. %as the output.
To further demonstrate the robustness of our method to noise, we follow~\cite{weder2020routedfusion} and add synthetic noise to each individual depth frame (denoted by the prefix `N-').
% To further demonstrate the capability of our differentiable renderer on optimizing both the pose and geometry, we add gaussian noise  with scales of 0.03m and 2$^\circ$ for the translation and rotation of the pose, respectively.
We additionally used the well-known {ICL-NUIM}~\cite{icl-nuim} public benchmark containing 4 scan trajectories for testing only, to demonstrate the generalizability of our method.

% \parahead{Data Generation}%
% To generate the input point cloud% for training
% , target SDF and target normal for training, we use a random subset of the Matterport3D~\cite{Matterport3D} RGB-D frames (50\% in our experiments) to construct input point cloud.
% \jhc{Is this true also for ICL-NUIM dataset?}\chx{ICL is for test, I add "for training" in the sentence.}
% The input point cloud is directly accumulated from \tj{posed} depth image and voxel down-sampling is then applied on the point cloud\jhc{Voxel down-sampling applied to what?}\chx{for point cloud}. Surface normal and sdf values are extracted from the full-scan ground-truth meshes provided by MatterPort3D.\jhc{What is 'groundtruth' meshes? Are they partial or full?}.\chx{It is not full every where but as full as possible} 

\parahead{Parameter Settings}%
Our \netname{} was trained and tested on a single Nvidia GeForce RTX 2080Ti GPU. 
%%%RRM Again, meaningless unless we know how big the inout scans are
The weights of the loss terms are empirically set to $\alpha = 0.1$, $\beta = 1$ and $\delta = 0.001$. 
We used the Adam optimizer with a learning rate of $0.001$. 
For efficient training, we uniformly split the input point cloud $\P$ into %3.2m$^3$ 
patches of size  $3.2\mathrm{m}\times 3.2\mathrm{m}\times 3.2\mathrm{m}$, although as a fully convolutional architecture, our pipeline could easily scale to the full scene during inferencing.
$\encnet$, $\unet$ and $\decnet$ have 4, 5, and  3 layers respectively.
\chx{With the scale of indoor scenes, the voxel size $b$ is set to 0.05m and the width of $\Omega_n$ is set to 2.5mm.}
Further details of our network structure are given in the supplementary material.

% \subsection{Experiment Settings}

% . We use traditional volumetric fusion method to construct the input of all the method that needs tsdf volumes or point clouds. This method is also a baseline.

\parahead{Baseline}%
Our method is compared to a full spectrum of methods, including those providing reconstruction from \chx{sequential depth frames}, \ie, RoutedFusion~\cite{weder2020routedfusion} (denoted {R-Fusion}) and {DI-Fusion}~\cite{huang2021di} using representations of either local implicit grid or a neural signed-distance volume.
We further consider methods operating on  fully-fused geometry, \ie, the 
convolutional occupancy network~\cite{peng2020convolutional} (denoted {ConvON}) is the state-of-art local implicit network for surface reconstruction considering global information, while {SPSG}~\cite{dai2021spsg} is the up-to-date scene completion approach that takes TSDF volumes as input. 
For methods that are cannot be trained on large-scale scenes, we used pre-trained weights obtained from synthetic datasets.
% For~\cite{huang2021di, peng2020convolutional, dai2020sg}, we change the input to the scene point cloud for a fair comparison.
% All the methods mentioned above claimed that their approach works well both on synthetic and real-world scenes, so we just use the pre-trained weight provided by the authors during evaluation. 
% \tjc{Are they also trained on the two datasets? If not, it seems not fair.}
% \jhc{This is one weakness in our experiments -- we have to either circumvent this or simply remove this sentence.}

\parahead{Metrics}%
We use root mean square error (RMSE), chamfer distance (CD), surface precision, recall, and F-score during evaluation. RMSE, CD, and surface precision mainly measure the accuracy of the reconstruction, surface recall mainly assesses  the degree of completeness, and F-score reflects both  accuracy and completeness.
All reconstruction results from different methods are converted to point clouds for comparisons.
RMSE and CD are measured \chx{in meters}, and for  %F-score, 
precision and recall, %and recall, 
a predicted or ground truth point is accepted if its distance to the closest ground truth or predicted point is smaller than $0.02$~m.
%%%RRM Again, absolute distance

\subsection{Comparisons to Other Methods}
\begin{figure*}
\includegraphics[width=\linewidth]{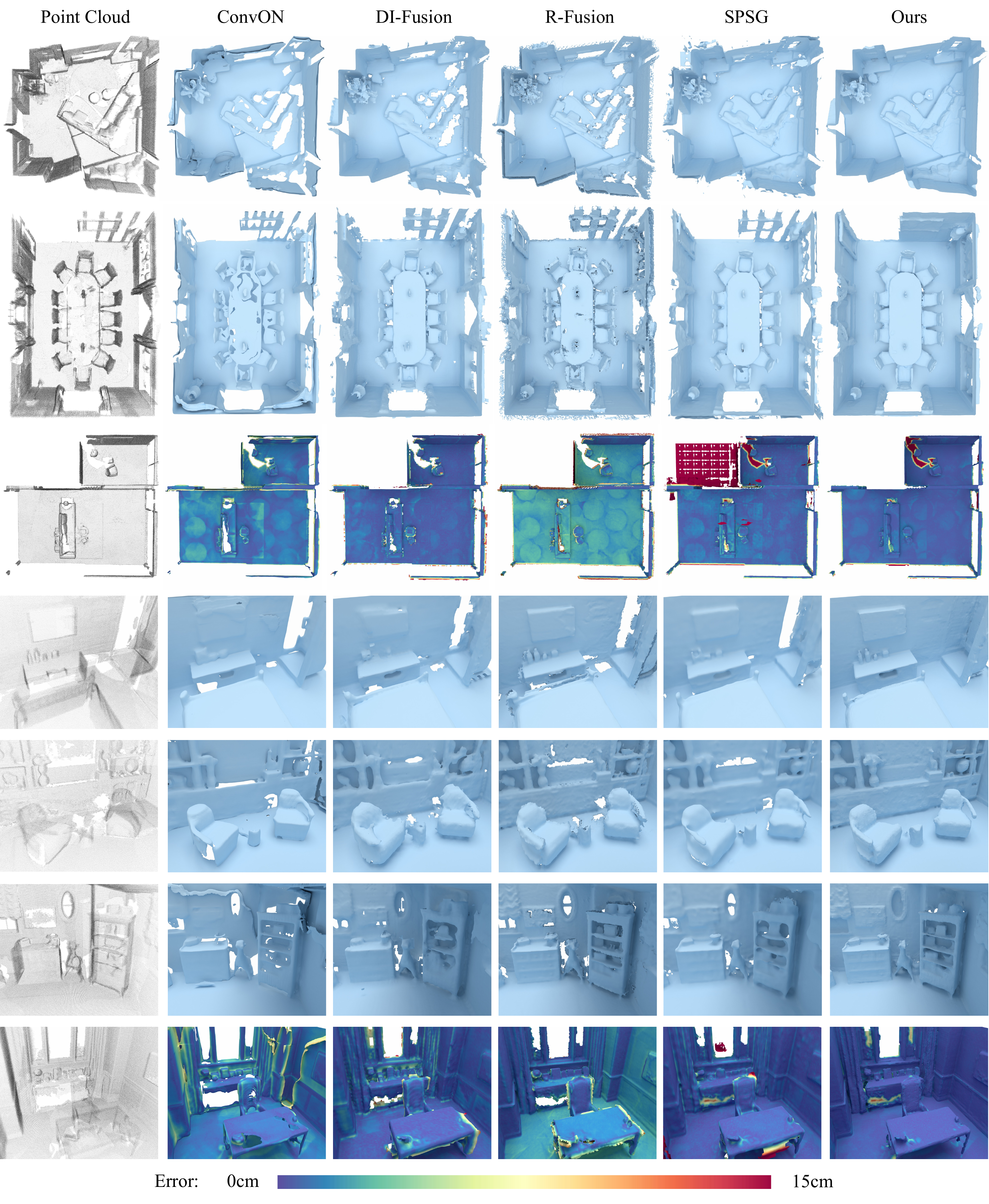}
\vspace{-2em}
\caption{\textbf{Visual comparison using N-Matterport3D.} Results show both  global views (part 1, top three rows) and  close-up views (part 2, bottom four rows). The last row in each part shows each method's per-point error, the distance between each reconstructed vertex and the corresponding closest ground truth point.}
\label{matter_qua}

\end{figure*}

% \parahead{Scene reconstruction with completion}%
\cref{matter_quan} shows that our proposed method works best according to all metrics, for the N-Matterport3D dataset. %F-score, recall and chamfer distance among all the baselines.
Qualitative results are presented in \cref{matter_qua}.  
The dense structure of ConvON makes it difficult for it to simultaneously capture local and global information from real-world datasets. 
R-Fusion and DI-Fusion only learn local geometric priors from the synthetic datasets. 
Specifically, although DI-Fusion fits local details with local implicit functions and achieves competitive performance, its lack of global information prevents it from completing missing regions. 
SPSG shows a capability for scene completion; however, limited by the discrete TSDF representation, the precision of the reconstructed surface is unsatisfactory. 
Our method learns global contextual information from local implicit grid by the convolutional neural network $\unet$, and thus can faithfully reconstruct local geometric details and recover many missing regions.
% TSDF-Fusion works well given ground-truth poses but lacks the ability for test-time refinement and adjustment if noise are present. \jhc{But it is still better than ours optimized?}

We further evaluate the generalizability of all  approaches using the ICL-NUIM dataset; quantitative results are given in \cref{icl_quan}. 
Remarkably, although our method is trained using panoramic scans as in~\cite{Matterport3D}, thanks to our effective learning scheme in 3D space, it generalizes well to hand-held trajectories whose geometric distributions are drastically different.%\tj{, while achieving a good balance between the reconstruction accuracy and completeness}.

\begin{table}[!tbp]
    \footnotesize
    \centering
    \setlength{\tabcolsep}{2.6pt}
    \begin{tabular}{@{}l|ccccc@{}}
    \toprule
    & \begin{tabular}[c]{@{}c@{}}RMSE $\downarrow$ \\ ($\times 10^{-3}$) \end{tabular} & \begin{tabular}[c]{@{}c@{}}CD $\downarrow$ \\ ($\times 10^{-3}$) \end{tabular} & \begin{tabular}[c]{@{}c@{}}F-Score $\uparrow$ \\ (\%) \end{tabular} & \begin{tabular}[c]{@{}c@{}}Precision $\uparrow$\\ (\%) \end{tabular} & \begin{tabular}[c]{@{}c@{}}Recall $\uparrow$\\ (\%) \end{tabular} \\ \midrule
    % TSDF~\cite{curless1996volumetric}         & 9.4       & 0.39 & { 88.79} & { 96.50}  & { 83.11} \\ \midrule
    SPSG~\cite{dai2021spsg}                   & 27.1      & 1.05    & { 80.12}      & { 76.03}      & { 85.21} \\
    ConvON~\cite{peng2020convolutional}       & 31.4      & 1.85    & { 62.34}      & { 52.32}      & { 78.45} \\
    R-Fusion~\cite{weder2020routedfusion}     & 24.6      & 0.98    & { 65.64}      & { 64.10}      & { 67.54} \\
    DI-Fusion~\cite{huang2021di}              & 20.9      & 1.14    & { 82.36}      & { 82.31}      & { 82.68} \\ \midrule
    Ours (w/o optim.)                                     & \underline{16.5}      & \underline{0.47}    & \underline{89.11}      & \underline{ 88.93}      & \underline{ 89.11} \\
    Ours                          & \textbf{16.2}      & \textbf{0.47} &  \textbf{89.23} & \textbf{ 89.23}  & \textbf{ 89.24} \\ \bottomrule
    \end{tabular}
    \caption{\textbf{Quantitative results using the N-Matterport3D dataset.} $\downarrow / \uparrow$: Lower / higher is better. \textbf{Bold} numbers indicate the best and \underline{underlined} numbers indicate the second best.}
    \label{matter_quan}
    \end{table}
%%%RRM Units are lacking, at least for RMSE: m? mm? what?
\begin{table}[!tbp]
    \footnotesize
    \centering
    \setlength{\tabcolsep}{2.6pt}
    \begin{tabular}{@{}l|ccccc@{}}
    \toprule
    & \begin{tabular}[c]{@{}c@{}}RMSE $\downarrow$ \\ ($\times 10^{-3}$) \end{tabular} & \begin{tabular}[c]{@{}c@{}}CD $\downarrow$ \\ ($\times 10^{-3}$) \end{tabular} & \begin{tabular}[c]{@{}c@{}}F-Score $\uparrow$ \\ (\%) \end{tabular} & \begin{tabular}[c]{@{}c@{}}Precision $\uparrow$\\ (\%) \end{tabular} & \begin{tabular}[c]{@{}c@{}}Recall $\uparrow$\\ (\%) \end{tabular} \\ \midrule
    SPSG~\cite{dai2021spsg}                   & 36.6      & 2.07 & 20.29 & 29.70  & 15.62 \\
    ConvON~\cite{peng2020convolutional}       & 42.3      & 3.55 & 13.81 & 18.46  & 11.15 \\
    R-Fusion~\cite{weder2020routedfusion}     & 40.9      & 2.75 & 14.56  & 22.20     & 11.07 \\
    DI-Fusion~\cite{huang2021di}              & \textbf{19.5}      & \textbf{1.32} & 22.14 & \underline{51.21}  & 14.23 \\ \midrule
    Ours (w/o optim.)                                    & 22.7      & 1.54 & \underline{23.89} &  51.02  & \underline{15.78}  \\
    Ours                          & \underline{22.1}       & \underline{1.46} &  \textbf{25.54}  &\textbf{53.55}   &\textbf{16.99}  \\ \bottomrule
    \end{tabular}
    \caption{\textbf{Quantitative results using the ICL-NUIM dataset.} See \cref{matter_quan} for explanation.}
    \vspace{-1em}
    \label{icl_quan}
    \end{table}
%%%RRM Units are lacking, at least for RMSE: m? mm? what?

\subsection{Ablation Study}
\label{ablation}

\begin{figure*}
\centering
\includegraphics[width=\linewidth]{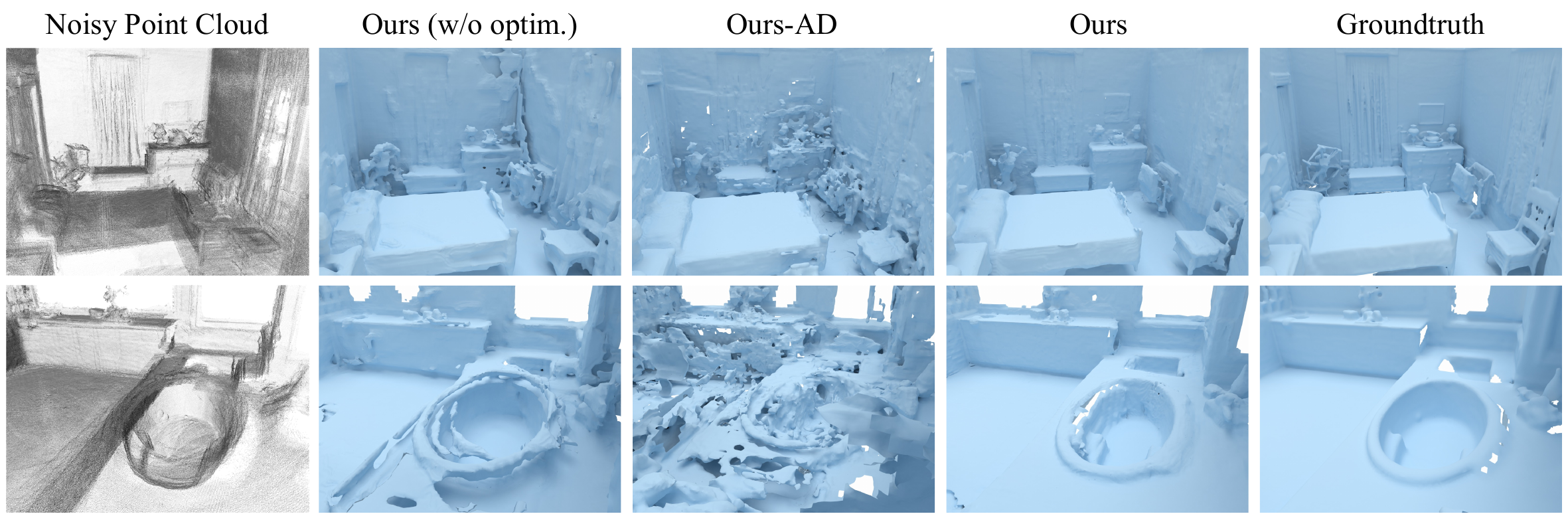}
\caption{\textbf{Inference-time refinement with differentiable rendering.} Our proposed method (Ours) effectively fixes the initial pose error (Ours w/o optim.) and produces a better reconstruction than the baseline (Ours-AD).}
\label{dr_ablation}
\vspace{-0.5em}
\end{figure*}

\parahead{Differentiable Rendering}
To demonstrate the capability of our differentiable renderer, we introduce a challenging scenario by adding  zero-mean Gaussian noise to the poses of frames from the N-Matterport3D dataset with a standard deviation of 3 cm
%%%RRM again, absolute distances
 and 2$^\circ$ for the translation and rotation, respectively.
Apart from direct comparisons with the version without differentiable rendering,
%%%RRM what unoptimized version? Of what? This needs an extra sentence of explanation.
 we verify the effectiveness of our implicit-differentiation-based gradient by replacing it by unrolled iterations obtained through automatic-differentiation~\cite{baydin2018automatic}, denoted by Ours-AD.
% As discussed in \cref{med:dr}, there are several differentiable approach, we implement two of them according to our sparse structure and show difference between two method. 
% One is our current setting, we call it Local-IDR(inspired by IDR\cite{yariv2020multiview}), and another one, illustrate in %TODO:add figure of dr
% , called Local-DIST(inspired by DIST\cite{liu2020dist}). 
% By experience, they all have the ability to optimize the feature volumn $\mathcal{V}$, however, there pose gradient of loss is totally different, so we provide the experiment of optimizing poses. 
As \cref{dr_ablation} shows, our renderer is able to denoise the input poses, reaching a higher reconstruction quality than its counterparts, the refinements of which are non-trivial due to the discrete TSDF representation used. 
% Based on traditional discreate tsdf representation, we can not easily finding a solution to jointly optimise the geometry and poses, resulting failure reconstruction as \cref{dr_ablation} shows.
Moreover, compared to Ours-AD, our full gradient optimization is also more effective, thanks to the accuracy and stability provided by the closed-form derivative computation.
% It's because that local-IDR leverage the information of $f(\rayc + t\rayd; \theta) \equiv 0$ and ensure that the first derivatives of pose are correct, so it works better than local-DIST when there are noisy poses. 
Our method also saves a considerable amount of optimization time and memory by avoiding propagating gradients through all  points along the ray. A detailed time and memory analysis of our differentiable rendering is given in the supplementary material.

%\begin{table}
%\caption{}
%\label{dr_timing}
%\end{table}

\begin{figure}
\centering
\includegraphics[width=\linewidth]{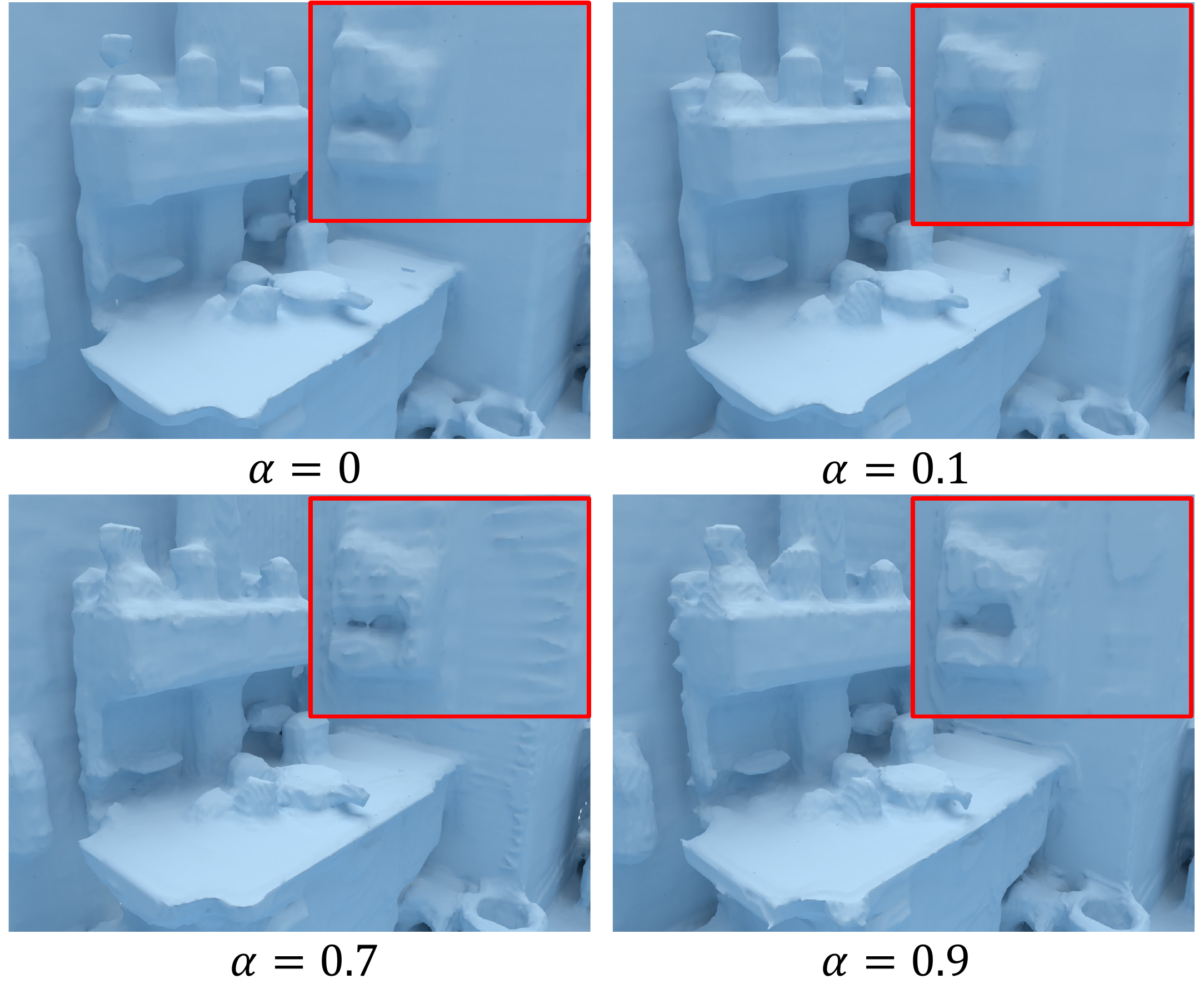}
\caption{\textbf{Effect of normal loss}. Reconstructions obtained by training with varying weights for $\mathcal{L}_\mathrm{norm}$, \ie, $\alpha$. Red boxes highlight differences.}
\vspace{-1em}
\label{loss_ab}

\end{figure}

\parahead{Weight of $\mathcal{L}_\mathrm{norm}$}
After fixing the gauge freedom of the weights for $\mathcal{L}_\mathrm{sdf}$ and $\mathcal{L}_\mathrm{struct}$ to 1, we show the effect of changing $\mathcal{L}_\mathrm{norm}$ in \cref{loss_ab}  by varying its weight $\alpha \in [0,1]$.
The addition of normal loss can effectively improve the precision of the reconstruction. 
However it only works when $\alpha$ is small, showing the importance of carefully choosing the weight parameter, especially in our setting with a small localized voxel size.
% This infers that, same as clamp sdf value, normal loss work in a different way when we try to fit a local geometry with very small voxel size.

% As \cref{med:train} demonstrated, there are three main loss item in the loss function and each of them has a loss weight, 
% we will analyze the effect of these losses as following. 
% $\mathcal{L}_\mathrm{sdf}$ is the main loss item, absolutely network won't work without this loss, so we just set the loss weight to 1 as default value. 
% $\mathcal{L}_\mathrm{struct}$ effects the predictors of $\phi_U$, when training, the predictor module is separate from \jhc{But the predictor is still linked to previous branches?}\chx{It surely effect previous branch, but this loss won't effect the parameter of sdf-decoder, which directly decide the result. By the way I can't find a better reason to explain why there is no ablation of structure loss.} the reconstruction module, so change the weight of $\mathcal{L}_\mathrm{struct}$ won't change the final reconstruction result, and we just fix the weight $\beta$ at 1. 
% As for $\mathcal{L}_\mathrm{norm}$, we illustrate  the effect of it . 

\begin{figure}[!t]
    \centering
\includegraphics[width=\linewidth]{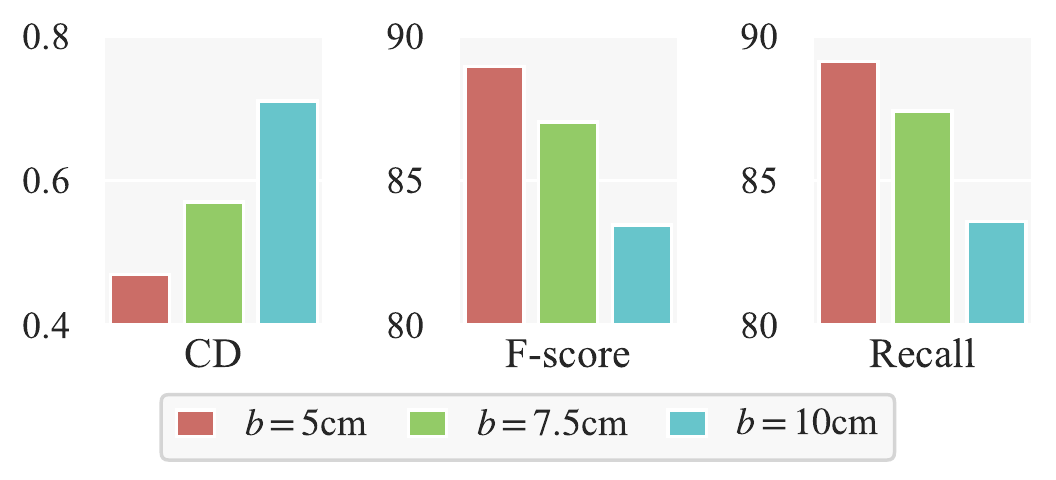}
\vspace{-2em}
\caption{\textbf{Performance for varying voxel sizes.} Our method works best with a small voxel size $b$.}
\vspace{-0.5em}
\label{vs_ab}
\end{figure}

\parahead{Voxel Size}%
\cref{vs_ab} shows how the voxel size $b$ (using 5~cm, 7.5~cm and 10~cm)
%%%RRM again, absolute values
 affects  reconstruction quality.
A smaller voxel size captures more details from the input and models the surface more accurately.
It also improves recall by avoiding mis-predicting large regions.
% The reconstruction quality increases with the decrease of voxel size no matter if the test time voxel size same as training time. 
Furthermore, empirically we find our method generalizes well across different voxel sizes: 
The test error with $b=7.5$~cm is stable even if trained using a different voxel size (test CD $\approx 0.019$ for both 5~cm and 10~cm).
Nevertheless, we recommend a larger voxel size during training to learn more complicated geometries for better generalization.

% When training and testing, voxel size is an important hyper parameter, we show that how voxel size affects the result in \cref{vs_ab}. There are five settings of voxel size, divide into two class: 1) same voxel size used in training and testing (s\_5, s\_7.5, s\_10), and 2) different voxel size used in training and testing (d\_1, d\_2). 

\subsection{Timing and Memory}
\begin{figure}[!t]
    \centering
    \includegraphics[width=\linewidth]{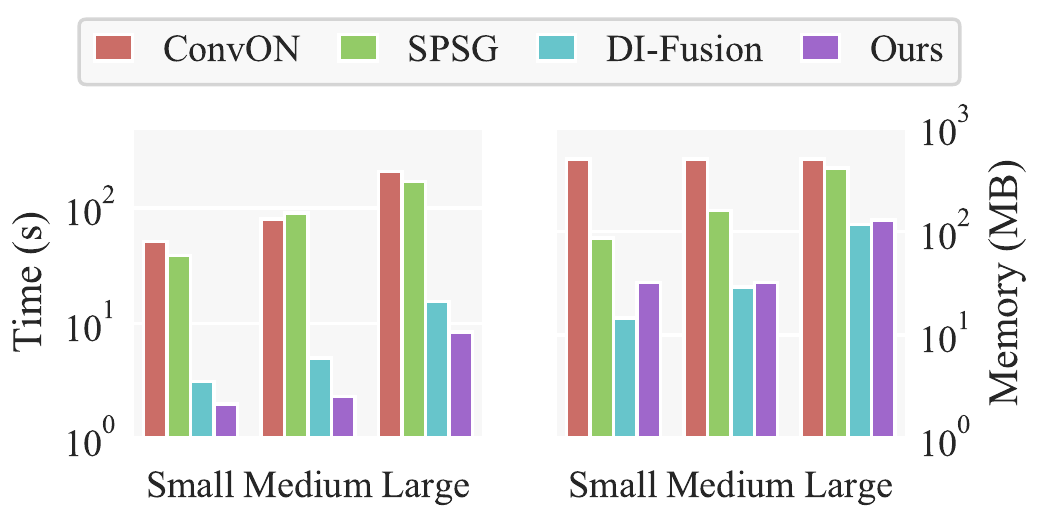}
\caption{\textbf{Run time and memory.} Results for different methods and  different input scene sizes.}
\label{time_me}
\vspace{-0.5em}
\end{figure}

Due to the differences in scene representation used by each approach, \chx{it is hard} to fairly compare the timing and memory consumption of the whole pipeline of each method. So we only compare the time to provide the discrete TSDF volume for a fair comparison. 
\cref{time_me} compares the inference time and memory footprint of the baselines for different scene sizes.
Thanks to the sparse feature volume, our method runs 10--50$\times$ faster than ConvON and SPSG, and is comparable in speed to DI-Fusion. However, as the scene gets larger, the time taken by DI-Fusion increases more rapidly than our method due to the difference in voxel interpolation strategy.
% As shown in , we pick three different size of the scene and record the inference  time of network. 
% In order to get neighborhood information, DI-Fusion need encode all points in 8-neighbor voxels in each voxel, result a bigger increment than our approach when scene become larger. 
As for memory cost, ConvON stays constant due to its sliding window inference scheme. 
SPSG maintains a dense discrete TSDF volume, so memory requirements grow drastically with  scene size.
Our method is memory-efficient due to its sparse representation and uses only marginally more memory than DI-Fusion while providing better reconstruction accuracy.
% However, in contrast to storing the feature in the center of voxels like DI-Fusion, our features are store in the corner of voxels to get smooth surface at the boundary of voxels, leading a little bigger memory use than DI-Fusion.

\subsection{Limitations and Discussion}
Our approach has three main limitations. 
Firstly, our network makes no use of  object-level priors, resulting in partially reconstructed objects even after completion. Training with semantic supervision may improve  completion performance.  
Secondly, reconstruction quality relies on a small voxel size that limits  further improvements in efficiency. This can be overcome with local implicit grid~\cite{Jiang_2020_CVPR}, which can learn local geometric priors from CAD models using large voxels with further optimization for real-world scenes. % shows that local geometric prior learned from CAD models with large voxels is powerful to reconstruct real-world scene based on optimization. So two stage training, first learn local prior from CAD models and second for global information on real-world, may be a good solution.
Thirdly, textures are not recovered by our method. Inspired by NeRF~\cite{mildenhall2020nerf}, training a neural radiance field together with SDF using differentiable rendering may be able to help  incorporate texture information into our pipeline.

%%%RRM Figure should say Medium not Middle

\section{Conclusions}

This paper has  introduced \name, a framework for large-scale scene reconstruction and completion using local implicit signed distance functions.
The key part of our method is a convolutional neural network that can learn global contextual information from local implicit grid, contributing to the completion of missing regions.
Together with our novel differentiable rendering strategy, we are able to generate an accurate and detailed reconstruction, while being fast and memory-efficient.
In the future, we hope to bridge the gap between large-scale geometric reconstruction and the use of object shape priors, as well as to incorporate color information into our pipeline, for better completion and reconstruction.

% \noindent\textbf{Acknowledgements.}
% {Placeholder}

% {\small
% \bibliographystyle{ieee_fullname.bst}
% \bibliography{main}
% }

{\small
\bibliographystyle{ieee_fullname.bst}
\bibliography{cvmbib}

\begin{thebibliography}{10}\itemsep=-1pt

\bibitem{azinovic2021neural}
Dejan Azinovi{\'c}, Ricardo Martin-Brualla, Dan~B Goldman, Matthias
  Nie{\ss}ner, and Justus Thies.
\newblock Neural rgb-d surface reconstruction.
\newblock {\em arXiv preprint arXiv:2104.04532}, 2021.

\bibitem{baydin2018automatic}
Atilim~Gunes Baydin, Barak~A Pearlmutter, Alexey~Andreyevich Radul, and
  Jeffrey~Mark Siskind.
\newblock Automatic differentiation in machine learning: a survey.
\newblock {\em Journal of machine learning research}, 18, 2018.

\bibitem{bovzivc2021transformerfusion}
Alja{\v{z}} Bo{\v{z}}i{\v{c}}, Pablo Palafox, Justus Thies, Angela Dai, and
  Matthias Nie{\ss}ner.
\newblock Transformerfusion: Monocular rgb scene reconstruction using
  transformers.
\newblock {\em arXiv preprint arXiv:2107.02191}, 2021.

\bibitem{Matterport3D}
Angel Chang, Angela Dai, Thomas Funkhouser, Maciej Halber, Matthias Niessner,
  Manolis Savva, Shuran Song, Andy Zeng, and Yinda Zhang.
\newblock Matterport3d: Learning from rgb-d data in indoor environments.
\newblock {\em International Conference on 3D Vision (3DV)}, 2017.

\bibitem{chen2021learning}
Yinbo Chen, Sifei Liu, and Xiaolong Wang.
\newblock Learning continuous image representation with local implicit image
  function.
\newblock In {\em Proceedings of the IEEE/CVF Conference on Computer Vision and
  Pattern Recognition}, pages 8628--8638, 2021.

\bibitem{chen2020bsp}
Zhiqin Chen, Andrea Tagliasacchi, and Hao Zhang.
\newblock Bsp-net: Generating compact meshes via binary space partitioning.
\newblock In {\em Proceedings of the IEEE/CVF Conference on Computer Vision and
  Pattern Recognition}, pages 45--54, 2020.

\bibitem{chibane2020implicit}
Julian Chibane, Thiemo Alldieck, and Gerard Pons-Moll.
\newblock Implicit functions in feature space for 3d shape reconstruction and
  completion.
\newblock In {\em Proceedings of the IEEE/CVF Conference on Computer Vision and
  Pattern Recognition}, pages 6970--6981, 2020.

\bibitem{curless1996volumetric}
Brian Curless and Marc Levoy.
\newblock A volumetric method for building complex models from range images.
\newblock In {\em Proceedings of the 23rd annual conference on Computer
  graphics and interactive techniques}, pages 303--312, 1996.

\bibitem{dai2020sg}
Angela Dai, Christian Diller, and Matthias Nie{\ss}ner.
\newblock Sg-nn: Sparse generative neural networks for self-supervised scene
  completion of rgb-d scans.
\newblock In {\em Proceedings of the IEEE/CVF Conference on Computer Vision and
  Pattern Recognition}, pages 849--858, 2020.

\bibitem{dai2017bundlefusion}
Angela Dai, Matthias Nie{\ss}ner, Michael Zollh{\"o}fer, Shahram Izadi, and
  Christian Theobalt.
\newblock Bundlefusion: Real-time globally consistent 3d reconstruction using
  on-the-fly surface reintegration.
\newblock {\em ACM Transactions on Graphics (ToG)}, 36(4):1, 2017.

\bibitem{dai2018scancomplete}
Angela Dai, Daniel Ritchie, Martin Bokeloh, Scott Reed, J{\"u}rgen Sturm, and
  Matthias Nie{\ss}ner.
\newblock Scancomplete: Large-scale scene completion and semantic segmentation
  for 3d scans.
\newblock In {\em Proceedings of the IEEE Conference on Computer Vision and
  Pattern Recognition}, pages 4578--4587, 2018.

\bibitem{dai2021spsg}
Angela Dai, Yawar Siddiqui, Justus Thies, Julien Valentin, and Matthias
  Nie{\ss}ner.
\newblock Spsg: Self-supervised photometric scene generation from rgb-d scans.
\newblock In {\em Proceedings of the IEEE/CVF Conference on Computer Vision and
  Pattern Recognition}, pages 1747--1756, 2021.

\bibitem{Genova_2020_CVPR}
Kyle Genova, Forrester Cole, Avneesh Sud, Aaron Sarna, and Thomas Funkhouser.
\newblock Local deep implicit functions for 3d shape.
\newblock In {\em Proceedings of the IEEE/CVF Conference on Computer Vision and
  Pattern Recognition (CVPR)}, June 2020.

\bibitem{Genova_2019_ICCV}
Kyle Genova, Forrester Cole, Daniel Vlasic, Aaron Sarna, William~T. Freeman,
  and Thomas Funkhouser.
\newblock Learning shape templates with structured implicit functions.
\newblock In {\em Proceedings of the IEEE/CVF International Conference on
  Computer Vision (ICCV)}, October 2019.

\bibitem{SubmanifoldSparseConvNet}
Benjamin Graham and Laurens van~der Maaten.
\newblock Submanifold sparse convolutional networks.
\newblock {\em arXiv preprint arXiv:1706.01307}, 2017.

\bibitem{icl-nuim}
A. Handa, T. Whelan, J.B. McDonald, and A.J. Davison.
\newblock A benchmark for {RGB-D} visual odometry, {3D} reconstruction and
  {SLAM}.
\newblock In {\em IEEE Intl. Conf. on Robotics and Automation, ICRA}, Hong
  Kong, China, May 2014.

\bibitem{hart1996sphere}
John~C Hart.
\newblock Sphere tracing: A geometric method for the antialiased ray tracing of
  implicit surfaces.
\newblock {\em The Visual Computer}, 12(10):527--545, 1996.

\bibitem{huang2021di}
Jiahui Huang, Shi-Sheng Huang, Haoxuan Song, and Shi-Min Hu.
\newblock Di-fusion: Online implicit 3d reconstruction with deep priors.
\newblock In {\em Proceedings of the IEEE/CVF Conference on Computer Vision and
  Pattern Recognition}, pages 8932--8941, 2021.

\bibitem{Jiang_2020_CVPR}
Chiyu~"Max" Jiang, Avneesh Sud, Ameesh Makadia, Jingwei Huang, Matthias
  Niessner, and Thomas Funkhouser.
\newblock Local implicit grid representations for 3d scenes.
\newblock In {\em Proceedings of the IEEE/CVF Conference on Computer Vision and
  Pattern Recognition (CVPR)}, June 2020.

\bibitem{kato2018neural}
Hiroharu Kato, Yoshitaka Ushiku, and Tatsuya Harada.
\newblock Neural 3d mesh renderer.
\newblock In {\em Proceedings of the IEEE conference on computer vision and
  pattern recognition}, pages 3907--3916, 2018.

\bibitem{liu2020neural}
Lingjie Liu, Jiatao Gu, Kyaw~Zaw Lin, Tat-Seng Chua, and Christian Theobalt.
\newblock Neural sparse voxel fields.
\newblock {\em NeurIPS}, 2020.

\bibitem{liu2019soft}
Shichen Liu, Tianye Li, Weikai Chen, and Hao Li.
\newblock Soft rasterizer: A differentiable renderer for image-based 3d
  reasoning.
\newblock In {\em Proceedings of the IEEE/CVF International Conference on
  Computer Vision}, pages 7708--7717, 2019.

\bibitem{liu2020dist}
Shaohui Liu, Yinda Zhang, Songyou Peng, Boxin Shi, Marc Pollefeys, and Zhaopeng
  Cui.
\newblock Dist: Rendering deep implicit signed distance function with
  differentiable sphere tracing.
\newblock In {\em Proceedings of the IEEE/CVF Conference on Computer Vision and
  Pattern Recognition}, pages 2019--2028, 2020.

\bibitem{lorensen1987marching}
William~E Lorensen and Harvey~E Cline.
\newblock Marching cubes: A high resolution 3d surface construction algorithm.
\newblock {\em ACM siggraph computer graphics}, 21(4):163--169, 1987.

\bibitem{mescheder2019occupancy}
Lars Mescheder, Michael Oechsle, Michael Niemeyer, Sebastian Nowozin, and
  Andreas Geiger.
\newblock Occupancy networks: Learning 3d reconstruction in function space.
\newblock In {\em Proceedings of the IEEE/CVF Conference on Computer Vision and
  Pattern Recognition}, pages 4460--4470, 2019.

\bibitem{mildenhall2020nerf}
Ben Mildenhall, Pratul~P Srinivasan, Matthew Tancik, Jonathan~T Barron, Ravi
  Ramamoorthi, and Ren Ng.
\newblock Nerf: Representing scenes as neural radiance fields for view
  synthesis.
\newblock In {\em European conference on computer vision}, pages 405--421.
  Springer, 2020.

\bibitem{newcombe2011kinectfusion}
Richard~A Newcombe, Shahram Izadi, Otmar Hilliges, David Molyneaux, David Kim,
  Andrew~J Davison, Pushmeet Kohi, Jamie Shotton, Steve Hodges, and Andrew
  Fitzgibbon.
\newblock Kinectfusion: Real-time dense surface mapping and tracking.
\newblock pages 127--136, 2011.

\bibitem{niemeyer2020differentiable}
Michael Niemeyer, Lars Mescheder, Michael Oechsle, and Andreas Geiger.
\newblock Differentiable volumetric rendering: Learning implicit 3d
  representations without 3d supervision.
\newblock In {\em Proceedings of the IEEE/CVF Conference on Computer Vision and
  Pattern Recognition}, pages 3504--3515, 2020.

\bibitem{oechsle2019texture}
Michael Oechsle, Lars Mescheder, Michael Niemeyer, Thilo Strauss, and Andreas
  Geiger.
\newblock Texture fields: Learning texture representations in function space.
\newblock In {\em Proceedings of the IEEE/CVF International Conference on
  Computer Vision}, pages 4531--4540, 2019.

\bibitem{oleynikova2017voxblox}
Helen Oleynikova, Zachary Taylor, Marius Fehr, Roland Siegwart, and Juan Nieto.
\newblock Voxblox: Incremental 3d euclidean signed distance fields for on-board
  mav planning.
\newblock In {\em 2017 IEEE/RSJ International Conference on Intelligent Robots
  and Systems (IROS)}, pages 1366--1373. IEEE, 2017.

\bibitem{park2019deepsdf}
Jeong~Joon Park, Peter Florence, Julian Straub, Richard Newcombe, and Steven
  Lovegrove.
\newblock Deepsdf: Learning continuous signed distance functions for shape
  representation.
\newblock In {\em Proceedings of the IEEE/CVF Conference on Computer Vision and
  Pattern Recognition}, pages 165--174, 2019.

\bibitem{peng2020convolutional}
Songyou Peng, Michael Niemeyer, Lars Mescheder, Marc Pollefeys, and Andreas
  Geiger.
\newblock Convolutional occupancy networks.
\newblock In {\em Computer Vision--ECCV 2020: 16th European Conference,
  Glasgow, UK, August 23--28, 2020, Proceedings, Part III 16}, pages 523--540.
  Springer, 2020.

\bibitem{qi2017pointnet}
Charles~R Qi, Hao Su, Kaichun Mo, and Leonidas~J Guibas.
\newblock Pointnet: Deep learning on point sets for 3d classification and
  segmentation.
\newblock In {\em Proceedings of the IEEE conference on computer vision and
  pattern recognition}, pages 652--660, 2017.

\bibitem{rosu2021neuralmvs}
Radu~Alexandru Rosu and Sven Behnke.
\newblock Neuralmvs: Bridging multi-view stereo and novel view synthesis.
\newblock {\em arXiv preprint arXiv:2108.03880}, 2021.

\bibitem{sitzmann2019siren}
Vincent Sitzmann, Julien~N.P. Martel, Alexander~W. Bergman, David~B. Lindell,
  and Gordon Wetzstein.
\newblock Implicit neural representations with periodic activation functions.
\newblock In {\em Proc. NeurIPS}, 2020.

\bibitem{song2018im2pano3d}
Shuran Song, Andy Zeng, Angel~X Chang, Manolis Savva, Silvio Savarese, and
  Thomas Funkhouser.
\newblock Im2pano3d: Extrapolating 360 structure and semantics beyond the field
  of view.
\newblock In {\em Proceedings of the IEEE Conference on Computer Vision and
  Pattern Recognition}, pages 3847--3856, 2018.

\bibitem{Sucar_2021_ICCV}
Edgar Sucar, Shikun Liu, Joseph Ortiz, and Andrew~J. Davison.
\newblock imap: Implicit mapping and positioning in real-time.
\newblock In {\em Proceedings of the IEEE/CVF International Conference on
  Computer Vision (ICCV)}, pages 6229--6238, October 2021.

\bibitem{sucar2021imap}
Edgar Sucar, Shikun Liu, Joseph Ortiz, and Andrew~J Davison.
\newblock imap: Implicit mapping and positioning in real-time.
\newblock In {\em Proceedings of the IEEE/CVF International Conference on
  Computer Vision}, pages 6229--6238, 2021.

\bibitem{takikawa2021neural}
Towaki Takikawa, Joey Litalien, Kangxue Yin, Karsten Kreis, Charles Loop, Derek
  Nowrouzezahrai, Alec Jacobson, Morgan McGuire, and Sanja Fidler.
\newblock Neural geometric level of detail: Real-time rendering with implicit
  3d shapes.
\newblock In {\em Proceedings of the IEEE/CVF Conference on Computer Vision and
  Pattern Recognition}, pages 11358--11367, 2021.

\bibitem{wang2020deep}
Peng-Shuai Wang, Yang Liu, and Xin Tong.
\newblock Deep octree-based cnns with output-guided skip connections for 3d
  shape and scene completion.
\newblock In {\em Proceedings of the IEEE/CVF Conference on Computer Vision and
  Pattern Recognition Workshops}, pages 266--267, 2020.

\bibitem{weder2020routedfusion}
Silvan Weder, Johannes Schonberger, Marc Pollefeys, and Martin~R Oswald.
\newblock Routedfusion: Learning real-time depth map fusion.
\newblock In {\em Proceedings of the IEEE/CVF Conference on Computer Vision and
  Pattern Recognition}, pages 4887--4897, 2020.

\bibitem{weder2021neuralfusion}
Silvan Weder, Johannes~L Schonberger, Marc Pollefeys, and Martin~R Oswald.
\newblock Neuralfusion: Online depth fusion in latent space.
\newblock In {\em Proceedings of the IEEE/CVF Conference on Computer Vision and
  Pattern Recognition}, pages 3162--3172, 2021.

\bibitem{whelan2015elasticfusion}
Thomas Whelan, Stefan Leutenegger, R Salas-Moreno, Ben Glocker, and Andrew
  Davison.
\newblock Elasticfusion: Dense slam without a pose graph.
\newblock 2015.

\bibitem{yariv2020multiview}
Lior Yariv, Yoni Kasten, Dror Moran, Meirav Galun, Matan Atzmon, Basri Ronen,
  and Yaron Lipman.
\newblock Multiview neural surface reconstruction by disentangling geometry and
  appearance.
\newblock {\em Advances in Neural Information Processing Systems}, 33, 2020.

\bibitem{yu2021plenoctrees}
Alex Yu, Ruilong Li, Matthew Tancik, Hao Li, Ren Ng, and Angjoo Kanazawa.
\newblock Plenoctrees for real-time rendering of neural radiance fields.
\newblock {\em arXiv preprint arXiv:2103.14024}, 2021.

\end{thebibliography}
}

% \newpage
% \setcounter{section}{0}
% \renewcommand\thesection{\Alph{section}}
% \input{sections/supplementary.tex}
% \newpage
% \input{symbols}
\newpage
\appendix

\setcounter{table}{0}
\setcounter{figure}{0}
\setcounter{equation}{0}
% \documentclass[10pt,onecolumn,letterpaper]{article}
% \usepackage{cvpr}

% % \cvprfinalcopy %

% \def\cvprPaperID{8008} %
% \def\httilde{\mbox{\tt\raisebox{-.5ex}{\symbol{126}}}}

% \ifcvprfinal\pagestyle{empty}\fi

% \title{Supplementary Material for \name{}: Convolutional Implicit Reconstruction and Completion for Large-scale Indoor Scene}

% \begin{document}
% \maketitle
\section{Network Architecture}
For our point encoder $\phi_\mathrm{E}$, we use a shared MLP model, which contains 4 layers including the input and output layers. The output feature size is set to 32.

\begin{figure*}[htbp]
% \begin{figure*}
\includegraphics[width=\linewidth]{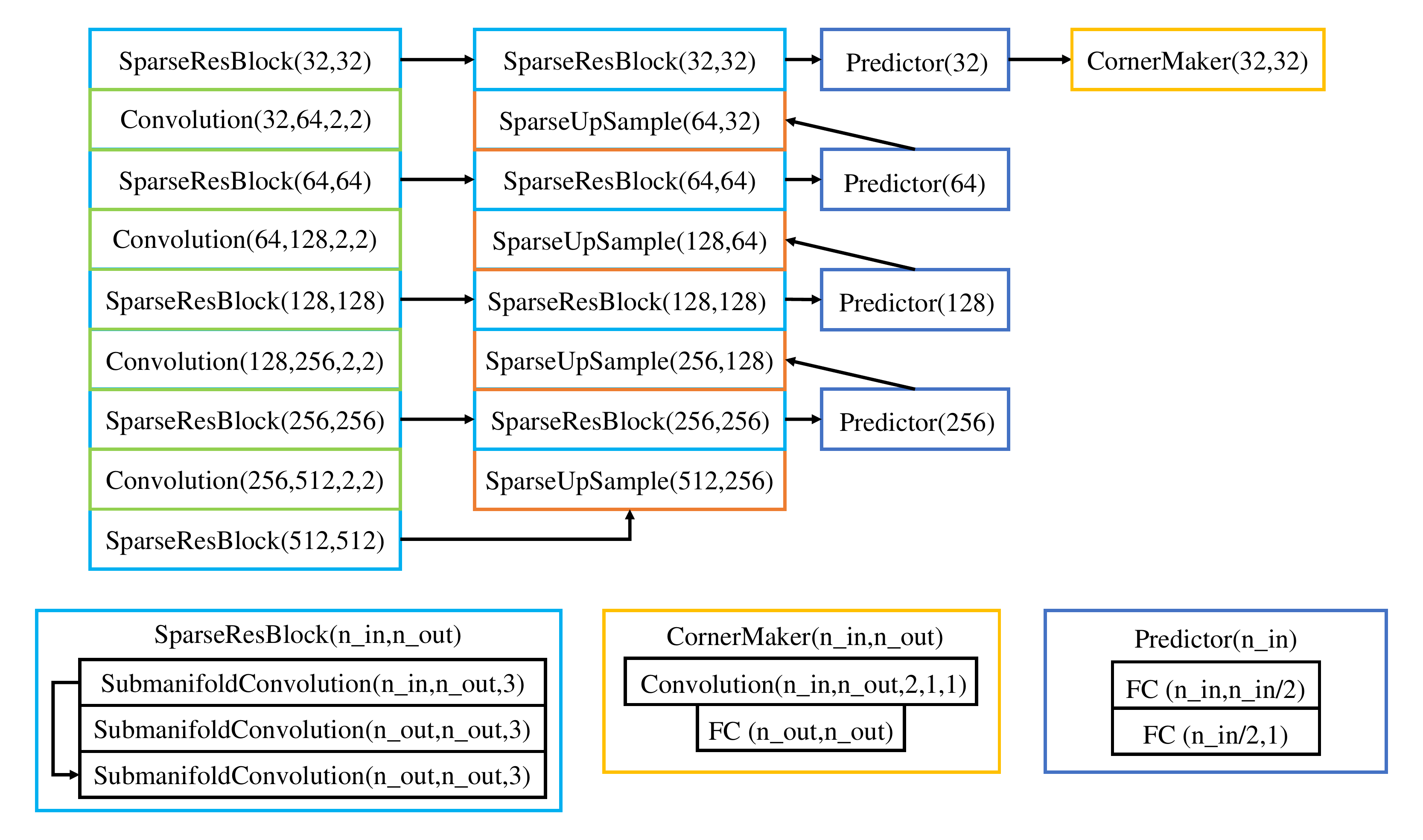}
\vspace{-1.5em}
\caption{\textbf{Detailed architecture of $\unet$.} Notice that all Fully-Connected (FC) layers receive sparse features as input and no dense tensor is built throughout the graph, significantly reducing the memory consumption.}
\vspace{-1em}
\label{fig_netar}
\end{figure*}

For our sparse U-Net $\phi_\mathrm{U}$, we illustrate it in \cref{fig_netar}. Convolution parameters are given in the format of (\texttt{n\_in}, \texttt{n\_out}, \texttt{kernel\_size}, \texttt{stride}, \texttt{padding}), where the stride and padding are set to 1 and 0 respectively as the default values. All convolutional layers and fully-connected layers except for the output layer are followed by instance normalization and LeakyRelu layers.  

For our SDF decoder $\phi_\mathrm{D}$, we use a small network which only contains 3 linear layers including the input and output layers and the channel sizes of the hidden layers are 64. Unlike other SDF decoders, the input latent vectors are not concatenated with the intermediate output of the network.

\section{Differentiable Renderer}
\subsection{Derivation}
In this section, we detail the procedure of implicit differentiation to obtain \cref{ep:dr} of the main paper. We denote the ray origin as $\rayc$, ray direction as $\rayd$, and the rendered depth as $t$, the hit point can be expressed as $\p = \rayc + t\rayd$.
Compute the total derivative of $f(\p,\theta)=0$ and we get:
\begin{equation}
\label{eq:total}
\frac{\partial{f}}{\partial{\p}}\frac{\partial\p}{\partial\rayc}\mathrm{d}\rayc + \frac{\partial{f}}{\partial{\p}}\frac{\partial\p}{\partial\rayd}\mathrm{d}\rayd + \frac{\partial{f}}{\partial{\p}}\frac{\partial\p}{\partial{t}}\mathrm{d}t + \frac{\partial{f}}{\partial\theta}\mathrm{d}\theta = 0,
\end{equation}
and according to $\p = \rayc + t\rayd$, we replace  ${\partial\p}/{\partial\rayc}$, ${\partial\p}/{\partial\rayd}$ and  ${\partial\p}/{\partial{t}}$ with $1$, $t$ and $\rayd$ respectively:

\begin{equation}
\label{eq:total_2}
\frac{\partial{f}}{\partial{\p}}\mathrm{d}\rayc + \frac{\partial{f}}{\partial{\p}}t\mathrm{d}\rayd + \langle\frac{\partial{f}}{\partial{\p}},\rayd\rangle\mathrm{d}t + \frac{\partial{f}}{\partial\theta}\mathrm{d}\theta = 0.
\end{equation}

To compute ${\partial{t}}/{\partial\theta}$, we ignore $\mathrm{d}\rayd$ and $\mathrm{d}\rayc$:
\begin{equation}
\langle\frac{\partial{f}}{\partial{\p}},\rayd\rangle\mathrm{d}t + \frac{\partial{f}}{\partial\theta}\mathrm{d}\theta = 0 \Rightarrow   \frac{\partial{t}}{\partial\theta} = -\langle\frac{\partial{f}}{\partial{\p}},\rayd\rangle^{-1}\frac{\partial{f}}{\partial\theta}.
\end{equation}

Similarly, we can compute the partial derivatives for $\rayc$ and $\rayd$:
\begin{align}
    \frac{\partial{t}}{\partial\rayc} = -\langle\frac{\partial{f}}{\partial{\p}},\rayd\rangle^{-1}\frac{\partial{f}}{\partial\p},\quad
\frac{\partial{t}}{\partial\rayd} = -t\langle\frac{\partial{f}}{\partial{\p}},\rayd\rangle^{-1}\frac{\partial{f}}{\partial\p}.
\end{align}

In our implementation, to satisfy the above partial derivatives, we construct the forward equation as:
\begin{equation}
t = t_0 + \frac{f(\p_0,\theta_0) - f(\p,\theta)}{\langle{\partial{f}}/{\partial{\p}}|_{\p = \p_0},\rayd_0\rangle},
\end{equation}
where $f(\p_0,\theta_0)$ means the SDF value provided by $\decnet$, and all the variables with subscript 0 are the constant values evaluated at the hit point.

\subsection{More Results}
\begin{figure}
\centering
\includegraphics[width=\linewidth]{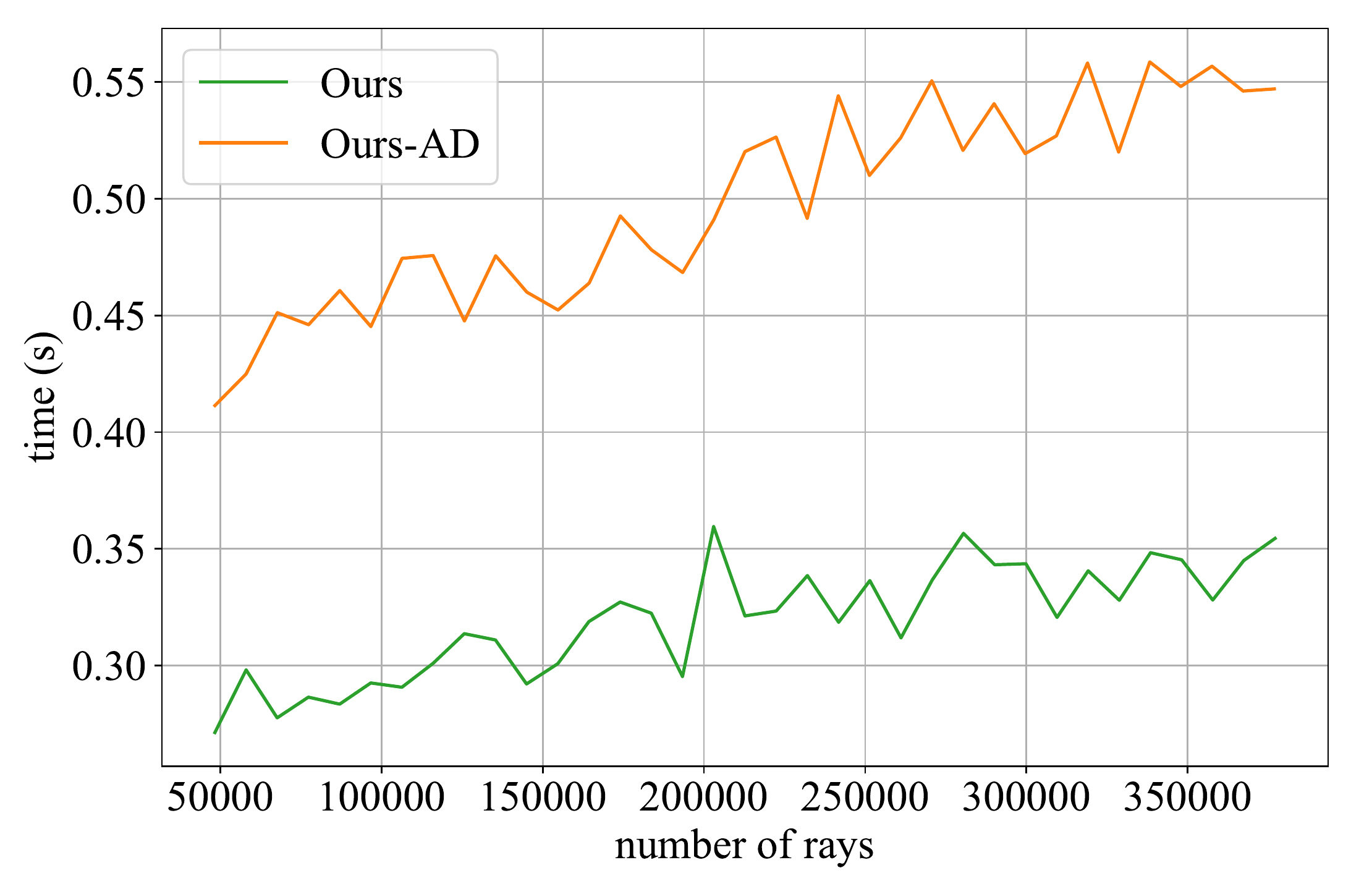}
\vspace{-2em}
\caption{\textbf{Speed comparison of two renderers.} We show that `Ours' is faster than `Ours-AD'. With the increasing number of rendered rays, rendering time grows slowly, implying that the performance bottleneck lies in the loop of the sphere tracing that is hard to be parallelized.}
\label{fig:dr_time}
\end{figure}

%
% \setlength{\columnsep}{20pt}
% \setlength{\intextsep}{0pt}
% \begin{wrapfigure}[18]{r}{\linewidth}
% \vspace{-1em}
% \begin{center}
% \includegraphics[width=\linewidth]{img/dr_time.pdf}
% \vspace{-2em}
% \caption{\textbf{Speed comparison of two renderers.} We show that `Ours' is faster than `Ours-AD'. With the increasing number of rendered rays, rendering time grows slowly, implying that the performance bottleneck lies in the loop of the sphere tracing that is hard to be parallelized.}
% \label{fig:dr_time}
% \end{center}
% \end{wrapfigure}
%

\begin{figure*}[!t]
\centering
\vspace{-0.5em}
\includegraphics[width=\linewidth]{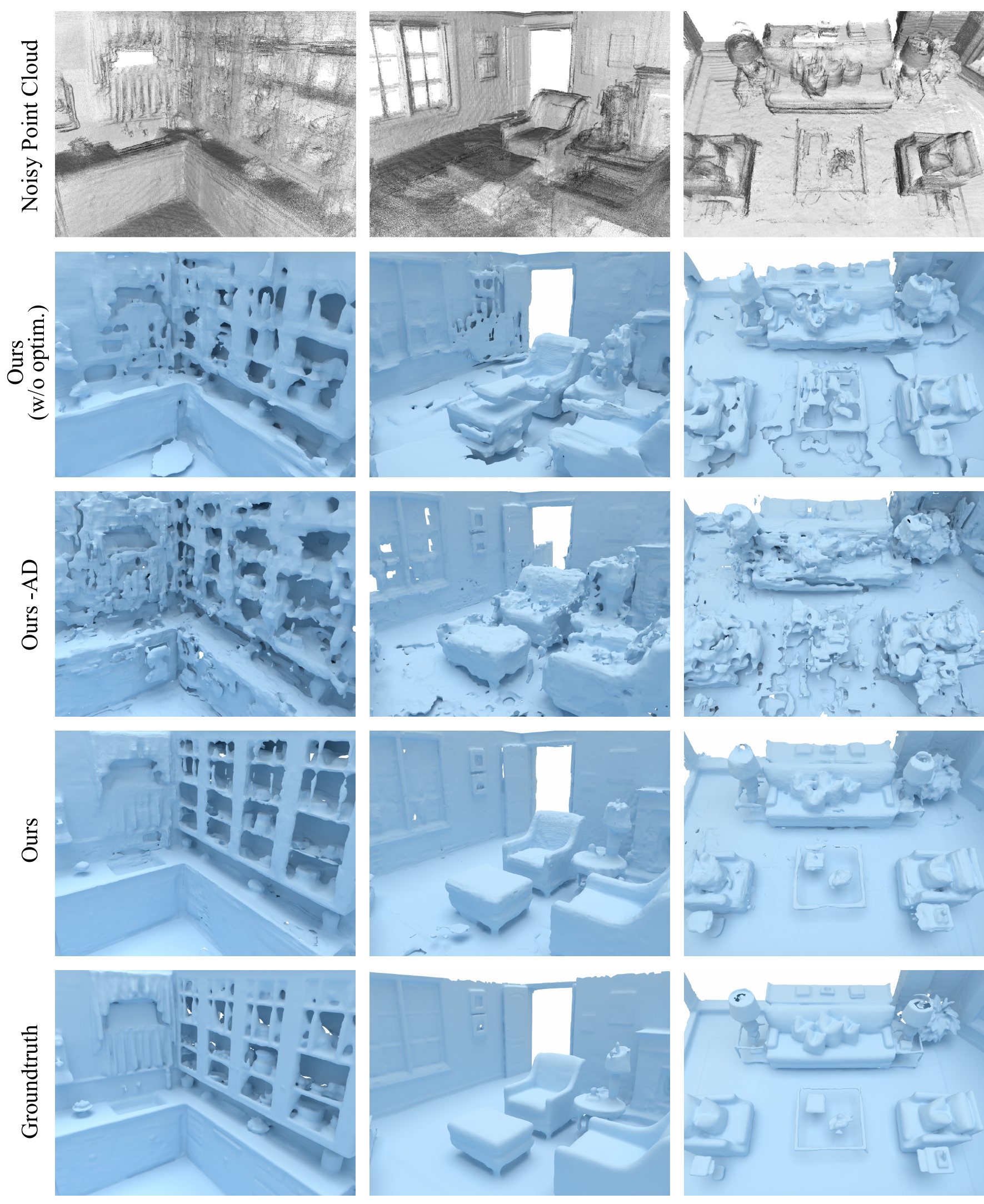}
\vspace{-1em}
\caption{\textbf{Qualitative results of differentiable renderer.} Our approach optimize geometry and poses jointly and generate fine-detailed mesh. Using implicit differentiation, our renderer provides more accurate gradient for poses than Ours-AD.}
\label{fig:more_dr}
\end{figure*}

We further demonstrate an alternative renderer using auto  differentiation provided by the deep learning framework, \ie Ours-AD, and show our differentiable renderer is faster and more accurate than Ours-AD in \cref{fig:dr_time} and \cref{fig:more_dr}.
For Ours-AD, rendered depth is given by:
\begin{equation}
    t = c + \sum^N_{i = 0}{f(\p_i,\theta)},
\end{equation}
where $c$ is the depth of ray-voxel intersect point and $\p_i$ is the $i$th point in sphere tracking procedure.

When rendering 300,000 rays, `Ours' takes about 1.3G GPU memory while `Ours-AD' takes about 4.6G. It is because `Ours-AD' stores all of the points $\p_i$ in the compute graph while `Ours' only stores the hit point.

% \begin{figure}
% \centering

% \end{figure}

\begin{figure*}[!t]
    \centering
    \includegraphics[width=0.95\linewidth]{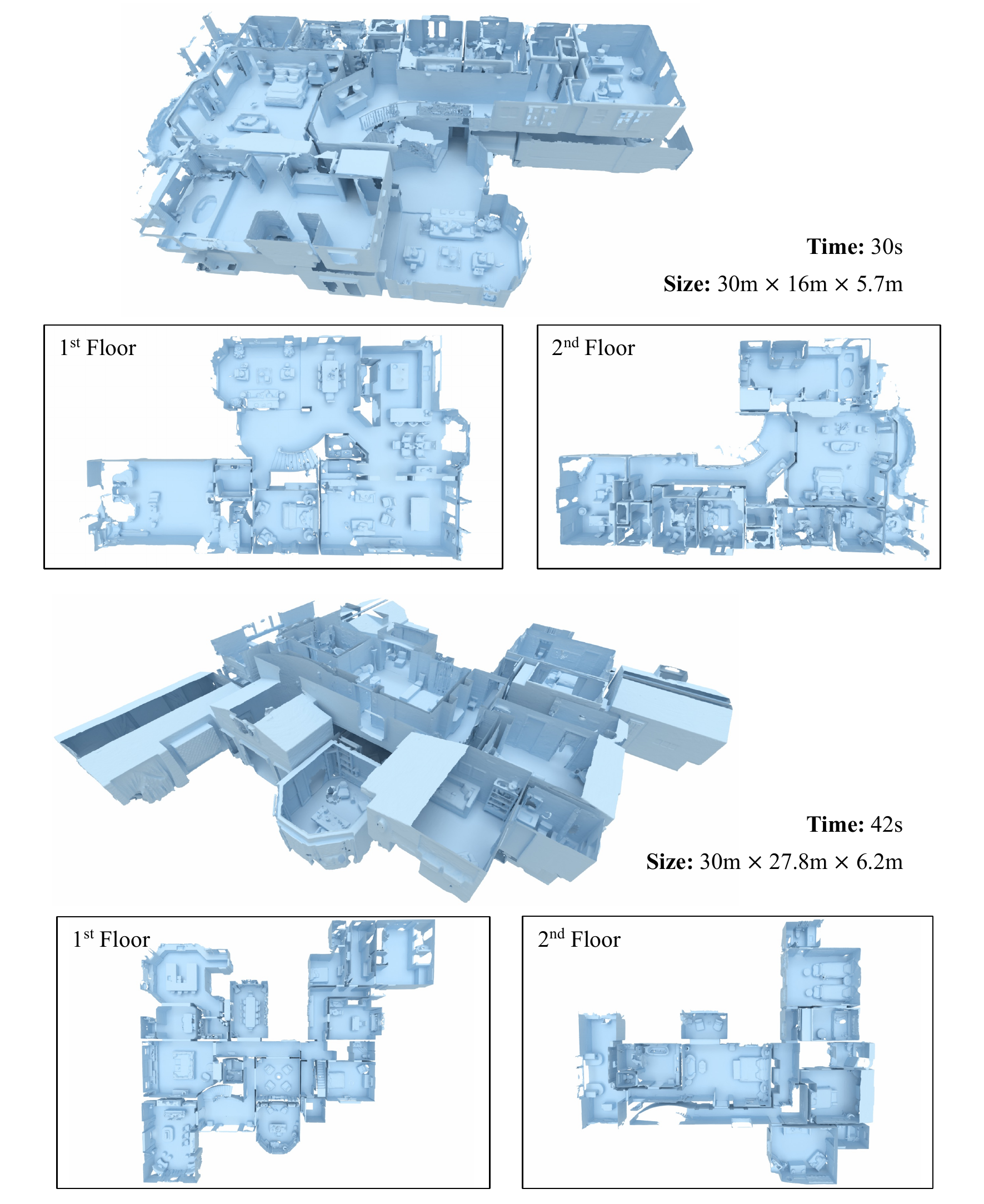}
    \caption{\textbf{Results on large scenes of Matterport3D.} We show two building-scale reconstructions from our method, with a single feed-forward pass. The sizes of buildings and the inference time are given on the right side of the figure and the subfigures in the bordered boxes show each floor.}
    \label{fig:large_scene}
\end{figure*}

\section{Reconstruction of Large Scenes}
As illustrated in \cref{fig:large_scene}, our method has the ability to reconstruct large scenes using a single feed-forward pass with a small run-time memory usage thanks to the sparse structure.

% \end{document}

\end{document}